\RequirePackage{fix-cm}
\documentclass[twocolumn]{svjour3}
\smartqed
\usepackage{amsmath,amsfonts}
\usepackage{algorithmic}
\usepackage{algorithm}
\usepackage{array}
\usepackage[caption=false,font=normalsize,labelfont=sf,textfont=sf]{subfig}
\usepackage{textcomp}
\usepackage{stfloats}
\usepackage{url}
\usepackage{verbatim}
\usepackage{graphicx}
\usepackage{graphicx}  % To include graphics
\usepackage{verbatim}  % To display text as-is
\usepackage{booktabs}  % To create better-looking table rules
\usepackage{multirow}  % To merge rows in a table
\usepackage{makecell}  % For flexible formatting within table cells
\usepackage{amssymb}  % To access additional math symbols
\usepackage{lipsum}  % To generate dummy text for testing or demonstration
\usepackage{colortbl}  % To color table cells
\usepackage{soul}  % For highlighting and strikethrough effects on text
\usepackage{microtype}  % For fine-tuning text layout
\usepackage{bm}  % For bold math symbols and vectors
\usepackage{makecell}  % Again for formatting within table cells
\usepackage{multirow}  % Again for merging rows in a table
\usepackage{soul}  % Again for text effects
\usepackage{enumitem}  % For customizing list environments
\usepackage{float}  % For better control of floating objects like figures and tables
\usepackage[colorlinks=true,citecolor=teal]{hyperref}
\usepackage{lineno}  % To add line numbers
\usepackage{array}  % For enhanced array and table formatting
\usepackage{xcolor}
\usepackage{natbib}
\usepackage{lmodern}
\usepackage{adjustbox} % for rotate table header

\newcolumntype{R}[2]{%
	>{\adjustbox{angle=#1,lap=1.3\width-(#2)}\bgroup}%
	l%
	<{\egroup}%
}

\definecolor{LightCyan}{rgb}{0.88,1,1}
\definecolor{better}{rgb}{0.19, 0.55, 0.91}
\definecolor{worse}{rgb}{0.82, 0.1, 0.26}

\newcommand{\major}[1]{\textcolor{black}{#1}} % First Revision
\newcommand{\revise}[1]{\textcolor{black}{#1}} % Second Revision
\newcommand{\minor}[1]{\textcolor{black}{#1}}

\journalname{International Journal of Computer Vision}

\begin{document}
	
\title{Is Contrastive Distillation Enough for Learning Comprehensive 3D Representations?}
\author{Yifan Zhang, Junhui Hou}

% Y. Zhang is with the School of Mechatronic Engineering and Automation, Shanghai University, Shanghai, China, and also with the Department of Computer Science, City University of Hong Kong, Hong Kong. E-mail: yfzhang@shu.edu.cn;
\institute{
	Yifan Zhang \at
	School of Mechatronic Engineering and Automation, Shanghai University, Shanghai, China\\
	Department of Computer Science, City University of Hong Kong, Hong Kong SAR, China\\
	\email{yfzhang@shu.edu.cn}\\
	Junhui Hou \at
	Department of Computer Science, City University of Hong Kong, Hong Kong SAR, China \\
	\email{jh.hou@cityu.edu.hk}
}

\date{Received: date / Accepted: date}

\maketitle
\begin{abstract}
	\revise{Cross-modal contrastive distillation has recently been explored for learning effective 3D representations. However, existing methods focus primarily on modality-shared features, neglecting the modality-specific features during the pre-training process, which leads to suboptimal representations. In this paper, we theoretically analyze the limitations of current contrastive methods for 3D representation learning and propose a new framework, namely CMCR (Cross-Modal Comprehensive Representation Learning), to address these shortcomings. Our approach improves upon traditional methods by better integrating both modality-shared and modality-specific features. Specifically, we introduce masked image modeling and occupancy estimation tasks to guide the network in learning more comprehensive modality-specific features.  Furthermore, we introduce a novel multi-modal unified codebook that learns an embedding space shared across different modalities. Besides, we propose geometry-enhanced masked image modeling to further boost 3D representation learning. Extensive experiments demonstrate that our method mitigates the challenges faced by traditional approaches and consistently outperforms existing image-to-LiDAR contrastive distillation methods in downstream tasks. Code will be available at~\href{https://github.com/Eaphan/CMCR}{\color{black}github.com/Eaphan/CMCR}.}
\end{abstract}

\section{Introduction}\label{sec:introduction}
LiDAR sensors have become essential tools for capturing detailed 3D information of the environment, playing a critical role in applications such as autonomous driving, robotics, and urban planning. The rich geometric data from LiDAR point clouds provide valuable spatial awareness that is difficult to achieve with traditional 2D sensors. However, processing these 3D point clouds often requires vast amounts of labeled data to train deep neural networks effectively. Annotating point cloud data, however, is both time-consuming and expensive, posing significant challenges to the scalability and practicality of 3D deep learning models~\citep{sautier2022slidr,chen2024building}. To address this issue, self-supervised learning has emerged as a promising solution, where networks are first trained on large volumes of unlabeled data~\citep{he2022masked,caron2021emerging}. This pre-training process enables the model to learn useful feature representations without the need for costly manual annotations. Once pre-trained, the network can be fine-tuned on smaller labeled datasets, greatly reducing the need for extensive labeling efforts and improving the efficiency of training~\citep{chen2020improved}. 

A widely used approach for learning 3D representations is contrastive pixel-to-point knowledge transfer, which leverages synchronized and calibrated images and point clouds~\citep{sautier2022slidr,mahmoud2023stslidr,liu2024seal,chen2024building,liao2024vlm2scene,puy24scalr,xu20254d}. The PPKT method~\citep{liu2021ppkt} allows a 3D model to benefit from the extensive knowledge encoded in a pre-trained 2D image backbone by using a pixel-to-point contrastive loss, with no need for labeled data for either the images or point clouds. Following this, SLidR~\citep{sautier2022slidr} introduces superpixels to group pixels and points that come from visually coherent regions, resulting in a more structured contrastive task. Building on these ideas, Seal~\citep{liu2024seal} leverages semantically rich superpixels generated by visual foundation models, incorporating temporal consistency regularization to enforce stability across point segments over time. Furthermore, CSC~\citep{chen2024building} investigates cross-scene semantic consistency for multi-modal 3D pre-training, aiming to ensure semantic coherence across all frames and scenes.

While contrastive learning methods have shown success in transferring knowledge between 2D and 3D modalities, they primarily focus on modality-shared information, which can limit their ability to fully capture the unique characteristics of each modality. These approaches emphasize aligning shared features across different modalities, but they often overlook the modality-specific details that could provide complementary insights. For instance, 3D point clouds contain rich spatial and geometric information, while 2D images encode fine-grained visual textures and color details. By concentrating on shared representations, these methods may miss out on leveraging the full potential of modality-specific features during pre-training, leading to suboptimal performance in downstream tasks that require a deeper understanding of each modality’s unique contributions~\citep{xu2013survey,liang2024factorized}.

In this work, we propose a novel approach that extends the traditional contrastive distillation paradigm by incorporating both modality-shared and modality-specific features. To achieve this, we design distinct heads that separately capture these features, driving the network to learn modality-specific information through tasks such as image reconstruction and 3D occupancy estimation. 
Additionally, we propose a new multi-modal unified codebook that aligns 2D and 3D features within a shared latent space. This codebook enables the model to focus on commonalities between modalities through shared features, while retaining the unique characteristics of each modality by utilizing separate heads for modality-specific features. By decoupling the shared and specific features, the codebook allows the model to effectively leverage both types of information and enhance performance on downstream tasks.
Furthermore, we leverage 3D features to assist in the reconstruction of masked image regions, and this process, in turn, enhances the learning of geometry-aware 3D representations.

\begin{figure}[t]
	\centering
	\includegraphics[width=0.4\textwidth]{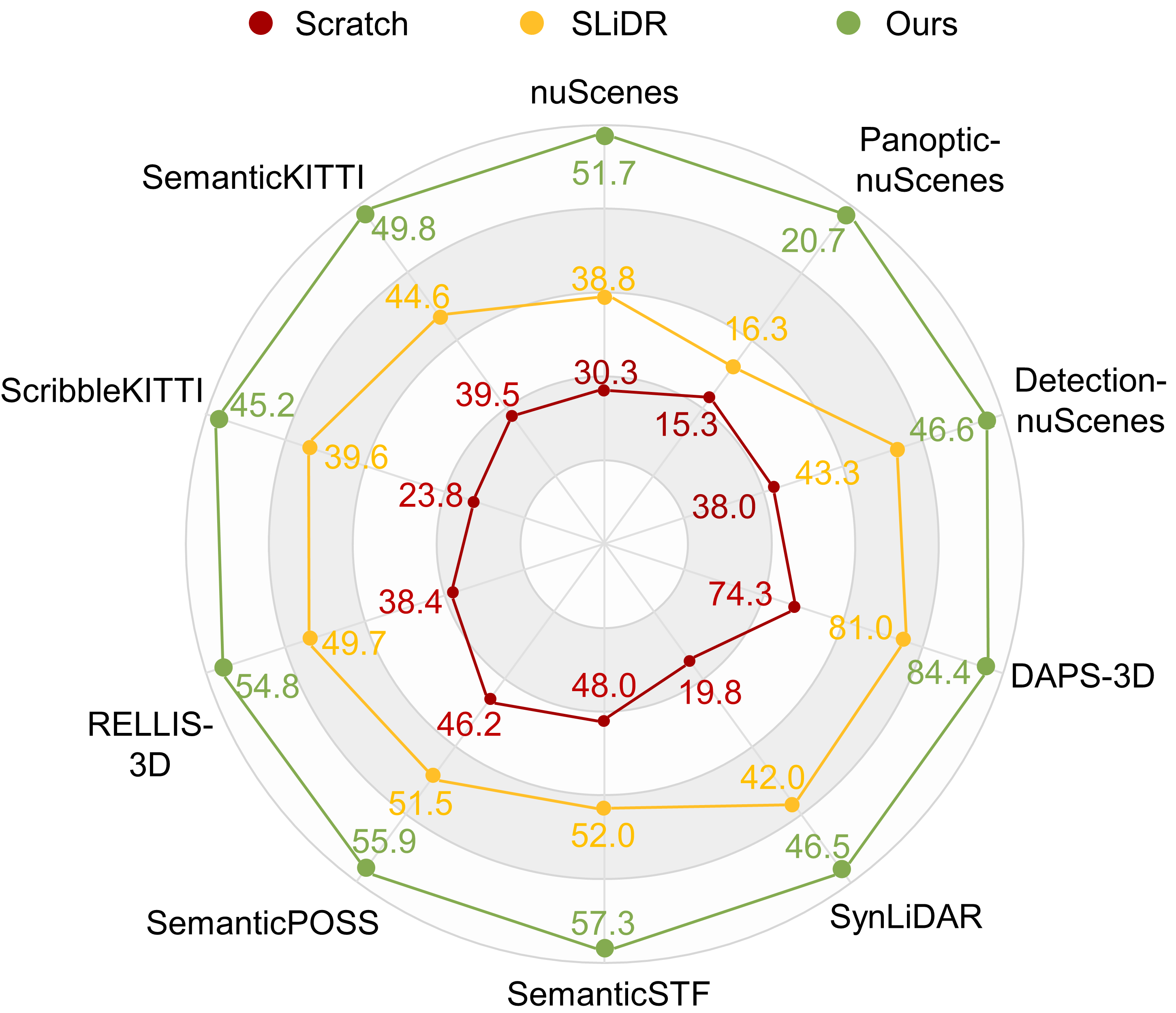} 
	\caption{
		Performance comparison of our method with scratch training and SLidR~\citep{sautier2022slidr} across multiple benchmarks. A larger covered area indicates superior overall performance.
	}
	\label{fig:radar}
\end{figure}

To assess the effectiveness of our method, we conduct extensive experiments and compare it with state-of-the-art approaches on several downstream tasks, including 3D semantic segmentation, object detection, and panoptic segmentation. The experimental results show that our method surpasses existing self-supervised learning techniques, as it demonstrates superior adaptability and performance across different tasks and datasets (see Fig.~\ref{fig:radar}). 

In summary, the primary contributions of this work are:

\begin{itemize}
	\item We provide a theoretical analysis of the limitations in current contrastive distillation methods for 3D representation learning.
	\item Building on the traditional contrastive distillation method, we introduce a new framework to jointly learn both modality-shared and modality-specific features.
	\item We propose a novel multi-modal unified codebook for learning a shared, modality-invariant embedding space.
	\item We propose the geometry-enhanced masked image modeling to enhance the 3D representation learning.
\end{itemize}

The remainder of this paper is organized as follows. Section~\ref{sec:related_work} reviews relevant prior work. Section~\ref{sec:analysis} presents a theoretical analysis of the limitations in current contrastive distillation methods. In Section~\ref{sec:method}, we describe the details of our proposed method. Section~\ref{sec:experiments} evaluates our method through experiments on three downstream tasks, along with ablation studies. Finally, we conclude the paper in Section~\ref{sec:conclusion}.

\section{Related Work}\label{sec:related_work}
This section provides an overview of existing research on 3D scene understanding, 3D representation learning, and codebooks, which are directly relevant to the core design of our approach.
\vspace{0.15cm}

\noindent\textbf{3D Scene Understanding.} Traditional methods for 3D scene understanding often rely on representations such as raw points~\citep{choe2022pointmixer,chen2022sasa}, voxels~\citep{puy2023using}, range views~\citep{kong2023rethinking,tian2022fully}, and multi-view fusion~\citep{fadadu2022multi,xu2021rpvnet} to capture environmental features. While effective, these approaches typically depend on large volumes of labeled data, which is both time-consuming and expensive to obtain, thereby hindering the scalability of 3D perception models~\citep{liu2022less}. To mitigate this issue, recent research has explored alternatives to reduce the reliance on fully annotated datasets. These include semi-supervised~\citep{kong2023lasermix,ho2024diffusion}, weakly-supervised~\citep{liu2023weakly,chibane2022box2mask}, self-supervised, and active learning~\citep{luo2023exploring,xie2023annotator} strategies, all of which aim to leverage minimal labeled data or automatically generate useful annotations to enhance model performance and scalability.
\vspace{0.15cm}

\noindent\textbf{3D Representation Learning.} Recent advancements in self-supervised learning for 3D point clouds have evolved alongside improvements in image-based methods~\citep{zhang2024hvdistill}. These include pretext tasks like predicting transformations or reconstructing point cloud parts~\citep{poursaeed2020self,sauder2019self}. \textit{Discriminative} methods focus on contrastive learning across different levels of representation (point, segment, region) to capture geometric and structural information~\citep{nunes2022segcontrast,yin2022proposalcontrast,chen20224dcontrast}. \textit{Temporal-consistency} methods leverage spatiotemporal correlations, aligning objects over time for robust representations~\citep{nunes2023TARL,huang2021STRL}. \textit{Reconstruction}\textit{-based} methods, such as those using Chamfer distance or surface reconstruction, aim to recover point cloud details from masked data. \revise{There is another type of method that conducts pre-training through \textit{cross-modal knowledge distillation}, as detailed below.}
\vspace{0.15cm}

\noindent\textbf{Image-to-LiDAR Contrastive Distillation.} 
\minor{Cross-modal distillation approaches utilize synchronized camera-LiDAR data for contrastive learning, transferring knowledge from 2D to 3D networks~\citep{sautier2022slidr,mahmoud2023stslidr,zhang2024fine,chen2024building,liao2024vlm2scene,puy24scalr,xu20254d}. The evolution of these methods reflects a continuous effort to enrich the alignment between modalities. Early foundational works focused on local region alignment, with~\cite{sautier2022slidr} introducing a superpixel-to-superpoint contrastive framework. To improve the robustness and semantic consistency of such regional alignments, subsequent methods built upon this foundation; for instance, ~\cite{mahmoud2023stslidr} integrated semantically tolerant constraints, while ~\cite{liu2024seal} incorporated spatial and temporal consistency regularization. Recent advancements have further diversified the alignment spaces and semantic targets. Structurally and temporally, ~\cite{zhang2024hvdistill} extended the contrastive domain to bird-eye views, and SuperFlow~\citep{xu20254d} introduced a 4D spatio-temporal framework utilizing flow-based contrast. Concurrently, approaches like OLIVINE and CSC leveraged the powerful priors of Visual Foundation Models (VFMs) to guide 3D representations via semantic prototypes~\citep{zhang2024fine,chen2024building}. Despite their distinct mechanisms and varied alignment domains---ranging from basic superpixels to complex 4D flows and VFM prototypes---these approaches share a fundamental commonality: they primarily focus on maximizing mutual information to align \textit{modality-shared} features. Consequently, they often overlook the \textit{modality-specific} details that provide complementary insights for 3D scene understanding. In this paper, we theoretically identify this shared limitation and propose a novel framework to capture both shared and specific representations for more comprehensive 3D learning.}
\vspace{0.15cm}

%\noindent\textbf{Image-to-LiDAR Contrastive Distillation.} Cross-modal distillation approaches utilize synchronized camera-LiDAR data to transfer knowledge from 2D to 3D networks~\citep{sautier2022slidr,mahmoud2023stslidr,zhang2024fine,chen2024building,liao2024vlm2scene,puy24scalr,xu20254d}. Recent advancements in this domain can be broadly categorized into three main directions based on their alignment strategies. First, \textbf{spatial and semantic consistency enhancements}: Methods like SLidR introduce superpixel-to-superpoint frameworks , which ST-SLidR improves via semantically tolerant constraints , and Seal extends through spatiotemporal regularization. Second, \textbf{integration of visual foundation models (VFMs)}: Approaches such as OLIVINE and CSC elevate 3D representations by utilizing VFM-guided semantic prototypes for higher-level cross-modal alignment~\citep{zhang2024fine,chen2024building}. Third, \textbf{structural and temporal expansions}: HVDistill explores representation alignment in bird-eye views~\citep{zhang2024hvdistill} , while SuperFlow introduces a 4D spatio-temporal framework utilizing flow-based contrast~\citep{xu20254d}. 

\noindent\textbf{Vector-Quantization and Codebook.}  
Vector quantization (VQ) is a technique originally introduced for image generation, where a large set of vectors is partitioned into clusters, each represented by a code vector from a codebook~\citep{van2017neural,peng2022beit}.
VQ was integrated into the autoencoder framework as VQ-VAE~\citep{van2017neural}, where it enables discrete latent representations by converting an image into a sequence of discrete codes and reconstructing it from these codes. This approach not only facilitates more compact and stable latent representations but also addresses issues like posterior collapse and variance instability that often affect traditional VAEs.
This technique has since been widely applied in various domains, including multimodal learning. Recent works have explored how to achieve a unified representation across multiple modalities by using a shared codebook~\citep{LiuJLROG22,xia2024achieving}. For example, Liu et al.~\citep{LiuJLROG22} proposed using a unified discrete space for aligning short videos and speech/text, addressing the codebook cold-start problem with a code warm-up strategy. \cite{chen2023revisiting} introduced FDT, which uses differentiable operations and can be trained end-to-end. In this work, we design a new multi-modal unified codebook to prevent the model from partitioning the codebook into modality-specific subspaces. Further technical details of our approach will be discussed in Section~\ref{sec:codebook}.

% Limitations of Existing Methods
% Review and Analysis of Current Methods
%\section{Challenges in Current Methods}
\section{Theoretical Analysis of Existing Approaches}\label{sec:analysis}
%In this section, we first

\subsection{Preliminary}
\noindent\textbf{Notation.} Define \( X^P = \{p_1, p_2, \dots, p_N|p_i \in \mathbb{R}^3\} \) as a point cloud of \( N \) points obtained from a LiDAR sensor, and \( X^{\mathcal{I}} = \{\mathcal{I}_c|c=1, \dots, N_{\mathrm{cam}}\} \) as a set of multi-view images captured by \( N_{\mathrm{cam}} \) synchronized cameras, where each image \( \mathcal{I}_c \in \mathbb{R}^{H \times W \times 3} \) has height \( H \) and width \( W \).

As a preliminary, we briefly review existing 2D-to-3D contrastive distillation techniques, particularly the point-to-pixel contrastive distillation framework from \cite{liu2021ppkt}, upon which we base our approach. Given point cloud and image data as inputs, we apply separate encoders for feature extraction. For the 3D point cloud, we use an encoder \( f_{\mathrm{3D}}(\cdot): \mathbb{R}^{N \times 3} \to \mathbb{R}^{N \times C_{\mathrm{3D}}} \) to generate per-point features of dimension \( C_{\mathrm{3D}} \). For images, we use an encoder \( f_{\mathrm{2D}}(\cdot): \mathbb{R}^{H \times W \times 3} \to \mathbb{R}^{H' \times W' \times C_{\mathrm{2D}}} \), which is initialized with weights from pre-trained image models. This framework supports knowledge transfer from the 2D domain to the 3D domain through contrastive learning. 

To compute the contrastive loss, we design trainable projection heads, \( h_{\mathrm{2D}} \) for 2D features and \( h_{\mathrm{3D}} \) for 3D features, which map the features to a common \( C \)-dimensional space. The 3D projection head \( h_{\mathrm{3D}} \) is a linear layer with \( \ell_2 \)-normalization, transforming 3D features into a normalized \( C \)-dimensional space. Similarly, the 2D projection head \( h_{\mathrm{2D}} \) consists of a 1×1 convolution followed by bilinear interpolation to adjust the spatial dimensions by a factor of 4, and it also applies \( \ell_2 \)-normalization.

Using the calibration matrix, we establish dense point-to-pixel correspondences \( \{ F^{P}_i, F^{\mathcal{I}}_i \}_{i=1}^{M} \), where \( F^{P}_i \) and \( F^{\mathcal{I}}_i \) are the features of matched points and pixels for the \( i \)-th pair, and \( M \) is the total number of valid pairs. Prior methods achieve cross-modal knowledge transfer by pulling positive pairs together and pushing negative pairs apart within the feature space, employing InfoNCE loss \citep{oord2018representation}. The point-pixel contrastive loss is defined as
\begin{equation}\label{eq:loss_point_pixel}
	\mathcal{L}_{\mathrm{NCE}} = - \frac{1}{M_s} \sum_{i=1}^{M_s} \log \left[\frac{\exp{(\langle F^{P}_i, F^{\mathcal{I}}_i \rangle / \tau)}}{\sum_{j=1}^{M_s} \exp{(\langle F^{P}_i, F^{\mathcal{I}}_j \rangle / \tau)}}\right],
\end{equation}
where \( \tau \) is a temperature parameter, \( M_s \) represents the number of sampled point-pixel pairs, and \( \langle \cdot, \cdot \rangle \) is the dot product used to measure feature similarity.

\begin{figure}[t]
	\centering
	\includegraphics[width=0.5\textwidth]{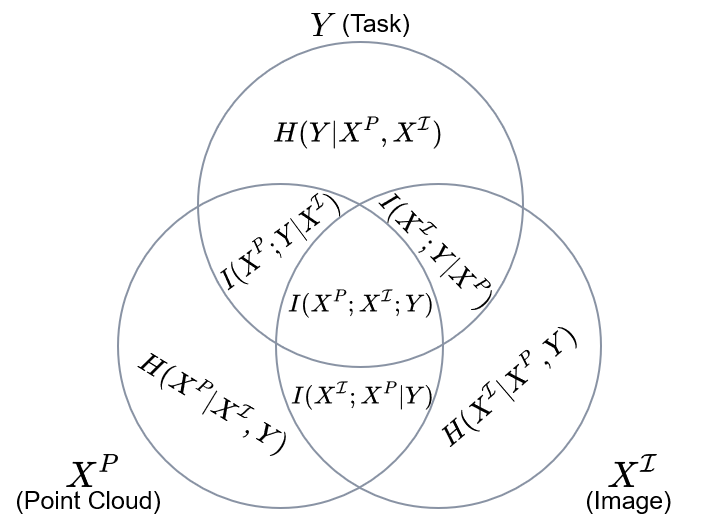} 
	\caption{
		  Depiction of the mutual information and entropy between the point cloud, image, and task-relevant information.
	}
	\label{fig:information}
\end{figure}

\subsection{Multi-view Non-redundancy Assumption}

In the context of point cloud data \( X^P \) and image data \( X^\mathcal{I} \), contrastive learning methods focus on maximizing the mutual information (MI) \( I(X^P; X^\mathcal{I}) \). These methods are based on the assumption that most task-relevant information is contained within the shared information between different views~\citep{sridharan2008information,xu2013survey}. However, because the background content in different views may vary, maximizing the MI across views can lead the encoder to prioritize the shared foreground information.

It is important to recognize that each view also contains unique task-relevant information that is specific to that view, which we refer to as modality-specific, task-relevant information. In other words, while shared information is essential, unique discriminative features in each view can also contribute significantly to the performance of downstream tasks. As illustrated in Fig.~\ref{fig:information}, solely focusing on modality-shared information may be insufficient. Including modality-specific task-relevant information can improve the general discriminative power of the learned representations.

To formalize these concepts, we define several types of information relevant to cross-modal representation learning. For the inputs \( X^P \) and \( X^\mathcal{I} \):
\begin{itemize}
	\item \( H(X^P) \) represents the entropy of \( X^P \).
	\item \( I(X^P; X^\mathcal{I}) \) denotes the mutual information between \( X^P \) and \( X^\mathcal{I} \), which we term as modality-shared information.
	\item \( I(X^P; Y|X^\mathcal{I}) \) and \( I(X^\mathcal{I}; Y|X^P) \) represent the task-relevant information that is unique to the point cloud and image inputs, respectively; we term this modality-specific information. Here, \( Y \) denotes the downstream task-relevant information.
\end{itemize}

It’s worth noting that in self-supervised learning (SSL), direct access to task-relevant information is unavailable. However, we hypothesize that an effective discriminative representation should incorporate both modality-shared and modality-specific information relevant to potential tasks. Specifically, the information for \( X^P \) related to \( Y \) can be decomposed as \( I(X^P; Y) = I(X^P; X^\mathcal{I}; Y) + I(X^P; Y|X^\mathcal{I}) \), which captures both the shared information across views and the unique task-relevant information in the point cloud.

\vspace{0.5em}
\noindent\textbf{Assumption 1.} There exists a constant \( \epsilon_u > 0 \) such that \( I(X^P; Y|X^\mathcal{I}) > \epsilon_u \).

This assumption suggests that cross-modal learning involves modality-specific task-relevant information, meaning that each modality (point cloud or image) contains unique information essential to the task that is not redundant between the two. Unlike traditional multi-view redundancy assumptions~\citep{sridharan2008information,xu2013survey}, where task-relevant information is assumed to be shared across views, this assumption allows for the existence of unique, non-overlapping information in each view. 

In practice, it is reasonable to assume \( I(X^P; Y|X^\mathcal{I}) > \epsilon_u \), as the point cloud \( X^P \) may contain crucial task-relevant information that the image \( X^\mathcal{I} \) cannot fully capture. For instance, the 3D spatial structure provided by the point cloud may be indispensable for certain tasks but is not entirely representable in a 2D image.

%\todo{In Figure ?, we illustrate this concept. Here, \( T \) denotes the task-relevant information.}

\subsection{Limitations of Contrastive Learning}
Current contrastive methods focus on maximizing the mutual information \( I(X^P; X^\mathcal{I}) \) between the two modalities without explicitly modeling modality-specific information. Typically, these approaches learn two representations as follows:

\begin{equation}\label{eq:nature_of_cl}
	F^{P}_{CL} = \mathop{\arg\max}_{F^{P}} I(F^{P}; X^\mathcal{I}),\quad F^{\mathcal{I}}_{CL} = \mathop{\arg\max}_{F^{\mathcal{I}}} I(F^{\mathcal{I}}; X^P).
\end{equation}

Here, $F^{P}_{CL}$ represents the encoding of the point cloud \( X^P \) and $F^{\mathcal{I}}_{CL}$ represents the encoding of the image \( X^\mathcal{I} \), both optimized by maximizing a lower bound on \( I(X^P; X^\mathcal{I}) \) using the Noise Contrastive Estimation (NCE) objective. In next analysis, we assume the contrastive distillation can achieve the optimal representation \( \{F^P_{CL}, F^{\mathcal{I}}_{CL}\} \) that satisfy Eq.~\eqref{eq:nature_of_cl} and \( I(F^P_{CL}; Y|F^\mathcal{I}_{CL}) = I(F^{\mathcal{I}}_{CL}; Y|F^P_{CL}) = 0 \).

However, under Assumption 1, standard contrastive learning methods face limitations. They primarily maximize a lower bound on the shared information \( I(X^P; X^\mathcal{I}) \), which provides only a limited training signal. This approach may struggle to capture task-relevant information that is unique to each modality. We formalize this intuition with the following observation:

\begin{theorem}[Suboptimality of Contrastive Distillation]\label{theorem:suboptimality}
	When modality-specific task-relevant information exists as described in Assumption 1, for the optimal learned representations \( \{F^P_{CL}, F^{\mathcal{I}}_{CL}\} \), we have:\\
\resizebox{0.48\textwidth}{!}{
	\begin{minipage}{\linewidth}
		\begin{align}
			I(F^P_{CL}; Y) &= I(X^P, X^\mathcal{I}; Y) - I(X^P; Y|X^\mathcal{I}) - I(X^\mathcal{I}; Y|X^P) \notag\\
			&= I(X^P; X^\mathcal{I}) - I(X^P; X^\mathcal{I}|Y) < I(X^P; Y).
		\end{align}
	\end{minipage}
}
\end{theorem}

This result implies that contrastive distillation, which only maximizes shared information, is suboptimal. It fails to capture the full task-relevant information present in each view, particularly the unique information specific to each modality, thereby limiting the discriminative power of the learned representations.

\noindent\textit{Proof of Theorem~\ref{theorem:suboptimality}:}
Since \( F^P_{CL} \) and \( F^{\mathcal{I}}_{CL} \) are the learned representations of \( X^P \) and \( X^\mathcal{I} \) that maximize the mutual information between them, we have:
\begin{equation}
	\resizebox{0.48\textwidth}{!}{
		$I(F^P_{CL}; Y) = I(X^P, X^\mathcal{I}; Y) - I(X^P; Y|X^\mathcal{I}) - I(X^\mathcal{I}; Y|X^P).$
	}
\end{equation}

This equation follows from the chain rule for mutual information and the fact that \( I(F^P_{CL}; Y|F^{\mathcal{I}}_{CL}) = I(F^{\mathcal{I}}_{CL}; Y|F^P_{CL}) = 0 \), meaning that all task-relevant information in \( Y \) is expected to be captured jointly by \( F^P_{CL} \) and \( F^{\mathcal{I}}_{CL} \).

By the chain rule of mutual information, we can decompose \( I(X^P, X^\mathcal{I}; Y) \) as follows:
\begin{equation}
	I(X^P, X^\mathcal{I}; Y) = I(X^P; Y) + I(X^\mathcal{I}; Y|X^P).
\end{equation}

Thus,
\begin{equation}
	I(F^P_{CL}; Y) = I(X^P; Y) - I(X^P; Y|X^\mathcal{I}) - I(X^\mathcal{I}; Y|X^P).
\end{equation}

According to Assumption 1, we have \( I(X^P; Y|X^\mathcal{I}) > \epsilon_u \). This means that there is modality-specific, task-relevant information in \( X^P \) that is not shared with \( X^\mathcal{I} \). Therefore, subtracting \( I(X^P; Y|X^\mathcal{I}) \) from \( I(X^P; Y) \) results in a reduction in the overall mutual information with \( Y \), capturing less than the total task-relevant information in \( X^P \).

Consequently, we have $I(F^P_{CL}; Y) < I(X^P; Y)$. It inequality demonstrates that maximizing only the mutual information between \( X^P \) and \( X^\mathcal{I} \) (as done in contrastive distillation) fails to capture the full task-relevant information available in \( X^P \), as it ignores modality-specific information unique to each modality. Hence, the learned representation \( F^P_{CL} \) is suboptimal for downstream tasks that require all task-relevant information.

\subsection{Generalization Ability of Learned Representations with Contrastive Methods}
Next, we provide a theoretical analysis of the generalization error for learned representations on a classification task, where \( Y \) is a categorical variable. We use the Bayes error rate as an example, representing the irreducible error (smallest generalization error) when predicting labels using any arbitrary classifier based on the learned representation.

Let \( P_e \) denote the Bayes error rate of the learned representations from the point cloud \( X^P \) and the multi-view images \( X^\mathcal{I} \), and let \( \hat{T} \) represent the estimated labels from our classifier.

Given the learned representations \( F^P \) from the point cloud and \( F^{\mathcal{I}} \) from the images, the Bayes error rate \( P_e \) can be bounded in terms of the mutual information between these representations and the task-relevant labels \( Y \).

\begin{theorem}\label{theorem:bayes_error}
	%The Bayes error rate \( P_e \), defined as the smallest achievable classification error with the optimal classifier 
	Given the learned representations $F_P$, the Bayes error rate \( P_e \) for a downstream classification task has an upper bound expressed as:
	\begin{equation}\label{eq:bayes_error}
		\hat{P}_e \leq 1 - \exp^{\left(- H(Y) + I(F^P; X^\mathcal{I}) + I(F^P; X^P|X^\mathcal{I}) - I(F^P; X^P|Y)\right)}.
	\end{equation}
\end{theorem}

\noindent\textbf{Remark.} 
Theorem~\ref{theorem:bayes_error} denotes that the error rate depends on the interplay between the mutual information terms:
1. \( I(F^P; X^\mathcal{I}) \): This term captures the modality-shared information between the learned representation \( F^P \) from the point cloud and the image view \( X^\mathcal{I} \). Maximizing this term helps ensure that \( F^P \) captures information that is relevant across both modalities, which can contribute to generalization. The term \( I(F^P; X^P|X^\mathcal{I}) \) is the modality-specific information in the point cloud representation \( F^P \) that is not shared with the image view \( X^\mathcal{I} \). Maximizing this term can allow the representation to capture unique, potentially task-relevant details specific to the point cloud view. This modality-specific information is crucial when the downstream task relies on unique aspects of the point cloud data that are not present in the images. The term \( I(F^P; X^P|Y) \) represents the task-relevant information in \( F^P \) specific to \( X^P \) when conditioned on the task label \( Y \). However, in a self-supervised learning (SSL) scenario, this quantity is unknown because labels are unavailable during training. Consequently, we cannot directly optimize for this term.

Since \( I(F^P; X^P|Y) \) is unknown and cannot be directly optimized in an SSL setup, we can focus on maximizing \( I(F^P; X^P|X^\mathcal{I}) \) and \( I(F^P; X^\mathcal{I}) \) to achieve a better representation. Doing so can enhance the generalization ability of the learned representation by capturing both shared and modality-specific information.

However, relying solely on contrastive distillation, which maximizes the mutual information between \( X^P \) and \( X^\mathcal{I} \)—is insufficient, as it primarily focuses on shared information \( I(X^P; X^\mathcal{I}) \) and may neglect modality-specific information \( I(F^P; X^P|X^\mathcal{I}) \). To address this limitation, we can incorporate reconstruction-based methods to better capture the modality-specific information.

By training the model to reconstruct the original point cloud \( X^P \) from the representation \( F^P \), we can enforce the representation to retain details specific to the point cloud that may not be present in the image view. This will effectively increase \( I(F^P; X^P|X^\mathcal{I}) \), making the representation more robust and improving the upper bound on the Bayes error rate.

In summary, to achieve a better generalization bound on the Bayes error rate, it is essential to focus not only on contrastive methods but also on methods that capture modality-specific information, such as reconstruction-based approaches. By maximizing both \( I(F^P; X^\mathcal{I}) \) and \( I(F^P; X^P|X^\mathcal{I}) \), we can create a richer representation that retains both shared and unique information across modalities, ultimately leading to improved performance on downstream tasks.

\vspace{0.5em}
\noindent\textit{Proof of Theorem~\ref{theorem:bayes_error}}:
Starting with the mutual information \( I(F^P; Y) \), we expand as follows:\\
\resizebox{0.49\textwidth}{!}{
	\begin{minipage}{\linewidth}
		\begin{align}
		I(F^P; Y) &= I(F^P; X^P) - I(F^P; X^P|Y) + I(F^P; Y|X^P) \notag\\
				&= I(F^P; X^P) - I(F^P; X^P|Y) \notag\\
				&= I(F^P; X^\mathcal{I}) - I(F^P; X^\mathcal{I}|X^P) + I(F^P; X^P|X^\mathcal{I}) \notag\\
				&  \quad  - I(F^P; X^P|Y) \notag\\
				&= I(F^P; X^\mathcal{I}) + I(F^P; X^P|X^\mathcal{I}) - I(F^P; X^P|Y).
		\end{align}
	\end{minipage}
}

Next, we use the inequality that relates \( P_e \) and \( H(Y|F^P) \) (from information-theoretic bounds~\citep{feder1994relations}):
\begin{equation}
	- \log(1 - P_e) \leq H(Y|F^P),
\end{equation}
where $H(Y|F^P)$ represents conditional entropy, i.e., the information can be obtained from $Y$ when $F^P$ is known.

By combining this with \( H(Y|F^P) = H(Y) - I(F^P; Y) \) and substituting \( I(F^P; Y) = I(F^P; X^\mathcal{I}) + I(F^P; X^P|X^\mathcal{I}) - I(F^P; X^P|Y) \), we obtain 
\begin{equation}
	\begin{aligned}
		\mathrm{log}(1 - P_e) \geq &- H(Y) + I(F^P;X^\mathcal{I}) + I(F^P;X^P|X^\mathcal{I}) \\ & - I(F^P;X^P|Y).
	\end{aligned}
\end{equation}

Rearranging, we find 
\begin{equation}
	P_e \leq \exp ^{\left(- H(Y) + I(F^P; X^\mathcal{I}) + I(F^P; X^P|X^\mathcal{I}) - I(F^P; X^P|Y)\right)}.
\end{equation}

\begin{figure*}[t]
	\centering
	\includegraphics[width=\textwidth]{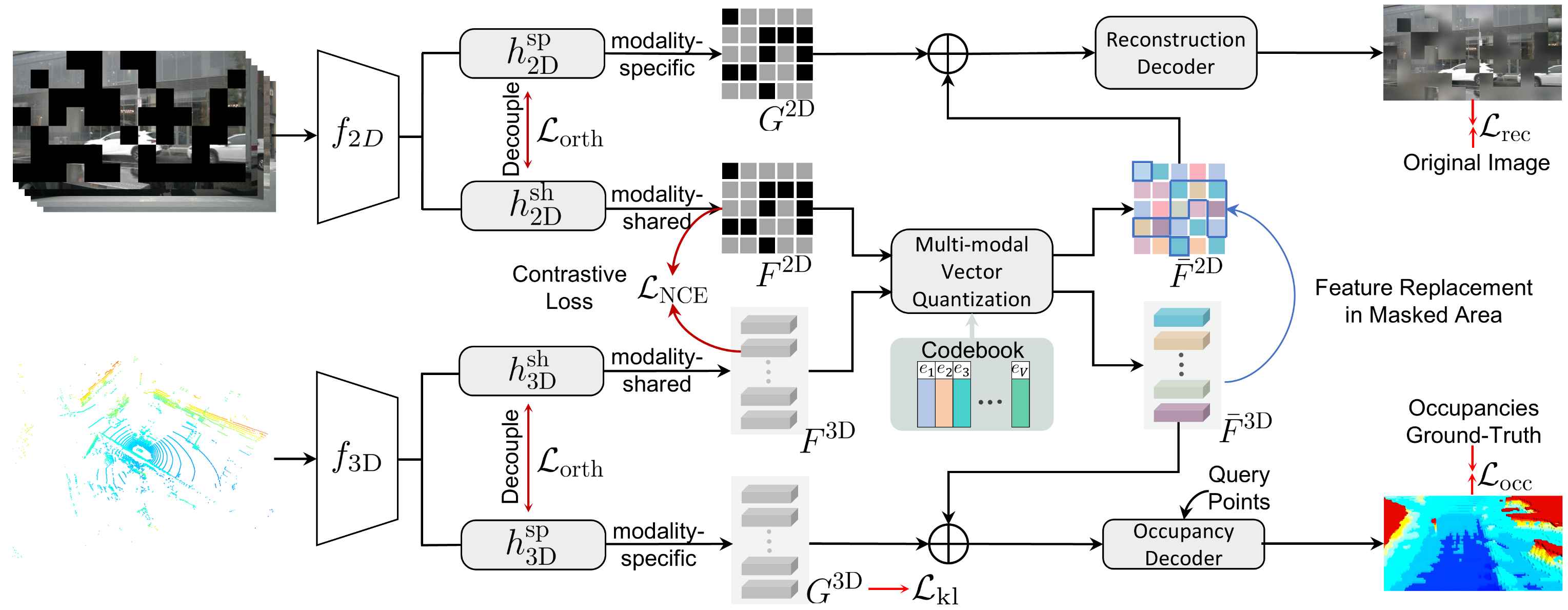} 
	\caption{
		\major{The overview of our proposed CMCR. The pipeline integrates both 2D image and 3D point cloud data to learn shared and modality-specific features. The model decouples features into two categories: modality-shared (denoted by \( F^{\mathrm{3D}} \) and \( F^{\mathrm{2D}} \)) and modality-specific (denoted by \( G^{\mathrm{3D}} \) and \( G^{\mathrm{2D}} \)). Contrastive learning is applied to modality-shared features, followed by vector quantization to map them to a unified latent space. The network is driven to learn modality-specific features with masked image restoration and occupancy estimation tasks.}
	}
	\label{fig:pipeline}
%	\vspace{-0.3cm}
\end{figure*}

\vspace{-0.3cm}
\section{Proposed Method}\label{sec:method}
\subsection{Overview}
%\noindent\textbf{Notation.} Let $P=\{p_1,p_2,...,p_N|p_i\in\mathbb{R}^3\}$ be a point cloud consisting of $N$ points collected by a LiDAR sensor, and $\mathcal{I} = \{I_c \|\ c=1,...,N_{\mathrm{cam}} \} $ multi-view images captured by $N_{\mathrm{cam}}$ synchronized cameras, where $I_c \in \mathbb{R}^{H \times W \times 3}$ is a single image with height $H$ and width $W$. 
Based on our deduction in Sec.~\ref{sec:analysis}, we propose a new 3D self-supervised learning method, namely CMCR (Cross-Modal Comprehensive Representation Learning), as depicted in Fig.~\ref{fig:pipeline}. For decoupling two types of features, we develop heads $\{h_{\mathrm{2D}}^{\mathrm{sh}}, h_{\mathrm{3D}}^{\mathrm{sh}}\}$ and $\{h_{\mathrm{2D}}^{\mathrm{sp}}, h_{\mathrm{3D}}^{\mathrm{sp}}\}$ to extract the modality-shared features $\{F^{\mathrm{3D}},F^{\mathrm{2D}}\}$ and modality-specific features $\{G^{\mathrm{3D}},G^{\mathrm{2D}}\}$, respectively. We perform contrastive learning based on the modality-shared features of point-pixel pairs. Then we perform vector quantization on these modality-shared features based on a multi-modal unified codebook module. Next, those embedding vectors and the modality-specific features are added together for masked image restoration and occupancy estimation.
These reconstruction tasks encourage the network to learn not only modality-shared features but also modality-specific features.

\vspace{0.5em}
\noindent\textbf{Choices of Modality-Specific Pretext Tasks.} \minor{1) For point clouds, we utilize occupancy estimation, which predicts whether queried spatial locations are occupied or empty. While Masked Autoencoders (MAE) are popular, applying voxel-based MAE to sparse outdoor scenes often necessitates masking inherently empty voxels to prevent the model from exploiting occupancy shortcuts. This forces the network to resolve whether a masked token represents a physically empty space or a masked object, introducing unnecessary ambiguity~\citep{hess2023masked}. In contrast, occupancy estimation circumvents this by providing a clear, direct supervisory signal for any spatial query, enabling the network to robustly learn 3D geometry and surface details. 2) MIM is selected for the image modality as it encourages the model to capture contextual information from visible regions, learning fine-grained texture, color, and spatial relationships unique to images. It also aligns 2D features with 3D representations for later fusion. Unlike Variational Autoencoders (VAEs), MIM does not require complex probabilistic modeling.}

\subsection{Multi-modal Unified Codebook}\label{sec:codebook}
To map the extracted modality-shared features into a unified latent space, we adopt a vector quantization (VQ) mechanism~\citep{van2017neural}. This approach discretizes the continuous representations into a finite set of codewords, ensuring consistent semantic alignment across different modalities. \major{Specifically, we define an embedding table \( E = \{e_1, e_2, \dots, e_V\} \in \mathbb{R}^{V \times C} \), where \( V \) is the size of the codebook, and \( C \) is the dimensionality of each codeword \( e_v \).}

Given the modality-shared features \( \{F^{\mathrm{3D}}, F^{\mathrm{2D}}\} \) extracted from the 3D point cloud and 2D image inputs, respectively, we pass these features through a discretization bottleneck. Each feature vector is then mapped to the nearest codeword in the codebook, as follows:
\begin{equation}
	\bar{F}^{\mathrm{M}}_i = F^{\mathrm{M}}_i + \mathrm{sg}(e_v - F^{\mathrm{M}}_i),
\end{equation}\vspace{-0.4cm}
\begin{equation}
	v = \underset{k \in \{1, \dots, V\}}{\mathrm{argmin}} \, ||F^{\mathrm{M}}_i - e_k||_2,
\end{equation}
where \( F^{\mathrm{M}}_i \) is a feature vector from modality $\mathrm{M} \in \{\mathrm{2D}, \mathrm{3D}\}$, and \( \mathrm{sg}(\cdot) \) denotes the stop-gradient operation. This ensures that the gradient flows only through the feature vector \( F^{\mathrm{M}}_i \) while treating the codeword \( e_v \) as a fixed target during backpropagation.

\vspace{0.5em}
\noindent\textbf{Codebook Update via Exponential Moving Average.} 
The codebook entries \( e_v \) are updated using an Exponential Moving Average (EMA) strategy to ensure stability and smooth updates during training. The EMA update for each codeword is computed as follows:
\begin{equation}
	\resizebox{0.49\textwidth}{!}{$
		e_v^{(t)} = \gamma e_v^{(t-1)} + \frac{1 - \gamma}{n^{\mathrm{2D}}_v(t) + n^{\mathrm{3D}}_v(t)} \left(\sum_{i=1}^{n^{\mathrm{2D}}_v(t)} F^{\mathrm{2D}}_i + \sum_{i=1}^{n^{\mathrm{3D}}_v(t)} F^{\mathrm{3D}}_i \right)$}
\end{equation}
where \( n_v^{\mathrm{2D}}(t) \) and \( n_v^{\mathrm{3D}}(t) \) are the counts of 2D and 3D features assigned to codeword \( e_v \) in the current batch, and \( \gamma \) is the decay rate for the moving average.

\begin{figure}[t]
	\centering
	\includegraphics[width=0.42\textwidth]{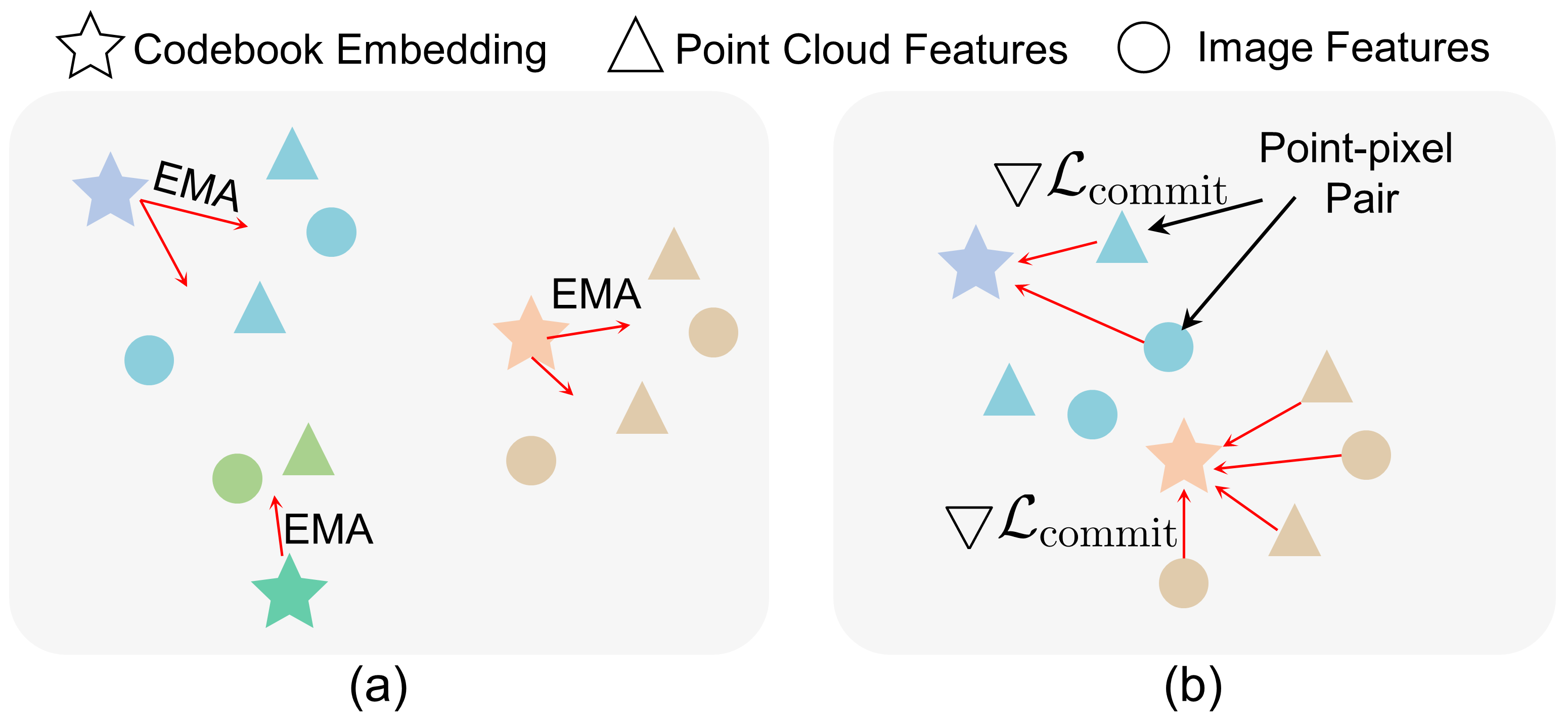} 
	\caption{
		(a) Illustration of the codebook update process using EMA.  (b) Depiction of the commitment loss mechanism, where 2D features are aligned with the codeword selected based on the corresponding 3D features.
	}
	\label{fig:codebook}
	\vspace{-0.3cm}
\end{figure}

\vspace{0.5em}
\noindent\textbf{Commitment Loss.} 
To ensure that the learned features are effectively quantized to the appropriate codewords, we introduce a novel commitment loss. The key challenge lies in the fact that the codebook, ideally shared across modalities, often partitions into modality-specific subspaces due to the distinct nature of fine-grained representations. To address this issue and ensure a unified embedding space that is invariant to modality, we promote consistency across modalities.

Specifically, we use 3D features to determine the codeword \( e_{v^*} \), and then align the 2D features to this selected codeword. This selective alignment ensures that both modalities converge towards a unified representation while maintaining the geometric structure inherent in the 3D modality. 

The commitment loss is formulated as follows:
\begin{equation}
    \resizebox{0.5\textwidth}{!}{
        $\mathcal{L}_{\text{commit}} = \frac{1}{n^{\mathrm{3D}}_{v}(t)} \left(\sum_{i=1}^{n^{\mathrm{3D}}_{v}(t)} ||F^{\mathrm{2D}}_i - \mathrm{sg}[e_{v^*}]||_2^2 + ||F^{\mathrm{3D}}_i - \mathrm{sg}[e_{v^*}]||_2^2\right),$
    }
\end{equation}
where \( v^* = \underset{k \in \{1, \dots, V\}}{\mathrm{argmin}} \, ||F^{\mathrm{3D}}_i - e_k||_2 \). 

By choosing the codeword based solely on 3D features, and enforcing 2D features to align with this codeword, we ensure that similar codewords are used for matching cross-modal pairs, i.e., a more consistent, modality-invariant cross-modal representation. 

\noindent\textbf{Remark.} The multi-modal unified codebook elevates cross-modal representation learning from simple pairwise alignment to structured semantic coherence. While conventional InfoNCE-based methods merely pull positive feature pairs closer in a continuous space, our codebook discretizes representations into shared, reusable semantic patterns. This mechanism not only enforces a more sophisticated structural alignment between image and point cloud modalities but also prevents modality collapse, ultimately yielding conceptually richer and more robust 3D representations.

\subsection{Geometry Enhanced Masked Image Modeling}

To effectively learn modality-specific features, we apply masked image modeling (MIM) within the image branch. Specifically, we apply a random masking strategy to the input images, obscuring certain patches to create a reconstruction task. When selecting point-pixel pairs for contrastive learning, we exclude pairs where the pixels fall within masked regions. This ensures that only unmasked, reliable features are used for contrastive alignment between the 2D and 3D modalities, resulting in more consistent cross-modal representations. The image reconstruction is performed using both modality-shared features \( \bar{F}^{\mathrm{2D}} \) and modality-specific features $G^{\mathrm{2D}}$ from the image branch, allowing the model to leverage both shared and unique information during the reconstruction process.

As we have aligned the 2D and 3D modality-shared features \( \bar{F}^{\mathrm{2D}} \) and \( \bar{F}^{\mathrm{3D}} \) through the unified codebook, we can further enhance the image reconstruction process by integrating 3D information. Specifically, in the masked regions of the image, we replace the missing modality-shared features \( \bar{F}^{\mathrm{2D}} \) with the corresponding aligned features from the 3D point cloud, \( \bar{F}^{\mathrm{3D}} \). This replacement leverages the structural information from the 3D modality to improve the quality of the reconstructed image, as the 3D features provide valuable spatial cues that are particularly helpful in regions where the image features are missing. Furthermore, by encouraging the image branch to leverage 3D features for reconstruction, the supervision also reinforces the learning of geometry-aware 3D representations.

\subsection{Occupancy Estimation}

We employ occupancy estimation as a type of 3D reconstruction task to drive the network toward learning modality-specific features in the 3D domain. Inspired by the occupancy estimation approach in ALSO~\citep{boulch2023ALSO}, we aim to predict the spatial occupancy around query points, where the model learns to classify these points as either ``occupied" or ``empty" based on the geometric structure of the environment.

\vspace{0.5em}
\noindent\textbf{Query Point Selection and Feature Extraction.}
To perform occupancy estimation, we randomly select query points within the 3D space. For each selected query point, we extract the surrounding point features from the combined representation of \( \bar{F}^{\mathrm{3D}} + G^{\mathrm{3D}} \), capturing both shared and specific details of the 3D environment. These features are then passed to an occupancy decoder, which predicts whether each query point lies in an occupied region of space or an empty region, effectively reconstructing the spatial structure around each point.

\vspace{0.5em}
\noindent\textbf{Decoder and Loss Function.}
Following ALSO~\citep{boulch2023ALSO}, we design the occupancy decoder as a multi-layer perceptron (MLP) that receives the feature vector of each query point and its relative position within the neighborhood. The decoder’s output is a binary classification for each query point, indicating occupancy status. We use a binary cross-entropy loss to supervise the occupancy predictions, where ground truth occupancies are derived based on sensor information and surface visibility, as described in ALSO~\citep{boulch2023ALSO}.

The reconstruction loss for occupancy estimation is defined as:
\begin{equation}
	\mathcal{L}_{\text{occ}} = -\frac{1}{|Q|} \sum_{q \in Q} o_q \log(\hat{o}_q) + (1 - o_q) \log(1 - \hat{o}_q),
\end{equation}
where \( Q \) is the set of query points, \( o_q \) is the ground truth occupancy, and \( \hat{o}_q \) is the predicted occupancy for each query point \( q \). This loss encourages the model to learn a precise occupancy map, capturing object-level and environmental details within the 3D space.

\vspace{0.5em}
\subsection{Overall Objective Function}
The overall objective function is designed to optimize multiple aspects of our model, balancing between cross-modal alignment, modality-specific learning, and reconstruction. The final loss function combines several components as follows:
\begin{equation}
	\mathcal{L}_{\text{total}} = \mathcal{L}_{\mathrm{NCE}} + \mathcal{L}_{\text{commit}} + \mathcal{L}_{\text{rec}} + \mathcal{L}_{\text{occ}} + \mathcal{L}_{\text{orth}} + \mathcal{L}_{\text{kl}},
\end{equation}
where reconstruction loss \( \mathcal{L}_{\text{rec}} \) is the loss for the MIM task, orthogonal loss \( \mathcal{L}_{\text{orth}} \) is designed to encourage independence between the modality-shared and modality-specific features within each modality,  and $\mathcal{L}_{\text{kl}}$ is the Kullback-Leibler (KL) Divergence loss developed for semantic consistency proposed in OLIVINE~\citep{zhang2024fine}. \major{The KL divergence loss is applied to the 3D features $G^{\mathrm{3D}}$.}

The loss term \( \mathcal{L}_{\text{orth}} \) ensures that the shared features capture information that is common across both modalities, while the specific features retain details unique to each modality, thus promoting disentangled and non-redundant representations. For each modality \( \mathrm{M} \in \{\mathrm{2D}, \mathrm{3D}\} \), the orthogonal loss \( \mathcal{L}_{\text{orth}} \) is defined to minimize the correlation between \( F^{\mathrm{M}} \) and \( G^{\mathrm{M}} \) by encouraging their inner product to be close to zero. This can be achieved by the following formulation:
\begin{equation}
	\mathcal{L}_{\text{orth}} = \sum_{\mathrm{M} \in \{\mathrm{2D}, \mathrm{3D}\}} \left\| \left( F^{\mathrm{M}} \right)^T G^{\mathrm{M}} \right\|_F^2,
\end{equation}
where \( \| \cdot \|_F \) denotes the Frobenius norm, which computes the squared sum of all entries in the matrix. 
\major{By minimizing the Frobenius norm of their inner product, the loss encourages these features to be uncorrelated, promoting disentangled and non-redundant representations.}
%The Frobenius norm of the inner product, $\| ( F^{\mathrm{M}} )^T G^{\mathrm{M}} \|_F^2$, quantifies the degree of correlation between the features. 
%\major{By minimizing this term, the model is encouraged to learn representations where $F^{\mathrm{M}}$ and $G^{\mathrm{M}}$ are orthogonal, meaning there is no overlap in the information they encode.}

\setlength{\tabcolsep}{4pt}
\begin{table}[t]
	\centering
	\addtolength{\extrarowheight}{\aboverulesep}
	\addtolength{\extrarowheight}{\belowrulesep}
	%	\setlength{\aboverulesep}{0pt}
	%	\setlength{\belowrulesep}{0pt}
	%	\caption{\todo{On nuScenes, CMCR is compared with current state-of-the-art methods in three downstream tasks with limited annotation. Obvious improvement in term of semantic segmentation, object detection, and panoptic segmentation could be found.}} 
	\caption{\revise{Comparison of CMCR with state-of-the-art methods on the nuScenes dataset across three downstream tasks with limited labeled data. The results show significant improvements in semantic segmentation, object detection, and panoptic segmentation at various ratios of available labeled data.}} % CMCR outperforms existing methods in all tasks, demonstrating its effectiveness in learning from limited annotations.
	\label{table:three_tasks_summary}
	\scalebox{0.75}{
		\vspace{-1.5em}
		\begin{tabular}{l|c|c}
			\Xhline{2\arrayrulewidth}
			\multirow{2}{*}{Method}	& \multicolumn{2}{c}{Semantic Segmentation (mIoU)}	\\
			\cline{2-3}
			{}	& 1\% 	& 5\%	\\ 
			\hline
			No Pre-training & 30.3	& 47.7	\\
			SLidR \citep{sautier2022slidr} & 38.2 & 52.2	\\
			ST-SLidR \citep{mahmoud2023stslidr}	& 40.7 & 54.6 \\   
%			TriCC \citep{pang2023unsupervised} & 41.2 & 54.1	\\ 
			Seal \citep{liu2024seal} & 45.8 & 55.6		\\ 
			CSC \citep{chen2024building} & 47.0 & 57.0		\\ 
			\revise{SuperFlow}~\citep{xu20254d}  & 49.9 & 60.7		\\ 
			\rowcolor{gray!20}
			Ours			& \textbf{51.7}			& \textbf{61.0}
			\\
			\hline
			\multirow{2}{*}{Method}	& \multicolumn{2}{c}{Object Detection (mAP / NDS)}
			\\
			\cline{2-3}
			{}		& 5\%	& 20\%
			\\
			\hline
			No Pre-training & 38.0 / 44.3	& 50.2 / 59.7
			\\
			SLidR~\citep{sautier2022slidr} & 43.3 / 52.4	& 50.4 / 59.9\\
			TriCC~\citep{pang2023unsupervised} & 44.6 / 54.4	& 50.9 / 61.3	\\
			CSC~\citep{chen2024building}	& 45.3 / 54.2 	& 51.9 / 61.3  		\\
			\rowcolor{gray!20}	Ours	& \textbf{46.6} / \textbf{55.2 }	& \textbf{52.7} / \textbf{62.0}  		\\
			\hline
			\multirow{2}{*}{Method}	& \multicolumn{2}{c}{Panoptic Segmentation (PQ / SQ / RQ)}	\\
			\cline{2-3}
			{}	& 1\% & 5\%	\\
			\hline
			No Pre-training & 15.3 / 62.6 / 20.4 & 20.9 / 73.4 / 26.5 \\
			SLidR~\citep{sautier2022slidr} & 16.3 / 65.7 / 21.4	& 21.6 / 73.5 / 27.1	\\
			CSC~\citep{chen2024building} &  19.3 / 74.5 / 24.6 & 23.1 / 76.9 / 28.5 	\\
			\revise{SuperFlow}~\citep{xu20254d} &  19.8 / 78.3 / 25.1 & 23.4 / 77.6 / 28.8 	\\
			\rowcolor{gray!20}	Ours & \bf 20.7 / 80.5 / 26.1 & \bf 24.0 / 78.5 / 29.3	\\
			\Xhline{2\arrayrulewidth}
		\end{tabular}
		\vspace{-1em}
	}
\end{table}
\setlength{\tabcolsep}{6pt}

\section{Experiments}\label{sec:experiments}
In this section, we present experimental results for three distinct 3D perception tasks, each addressed using a well-established 3D backbone relevant to its domain. Specifically, we evaluate semantic segmentation with MinkUNet~\citep{choy20194d} in Sec.~\ref{sec:exp_semantic_segmentation}, object detection using VoxelNet~\citep{zhou2018voxelnet} in Sec.~\ref{sec:exp_detection}, and panoptic segmentation with Cylinder3D~\citep{zhu2021cylindrical} in Sec.~\ref{sec:exp_panoptic}. To provide a thorough comparison of CMCR against existing approaches across these tasks, particularly with limited labeled data, we summarize the results in Table~\ref{table:three_tasks_summary}. Furthermore, we assess the contribution of each model component in Sec.~\ref{sec:ablation_study}. Due to space constraints, additional visualizations and experiments are available in the appendix.

\vspace{-0.5em}
%\subsection{Dataset and Evaluation Metric}\label{sec:dataset}
\subsection{Implementation Details}\label{sec:implementation_details}
\noindent\textbf{Datasets.} We pre-train all three models on the nuScenes dataset~\citep{caesar2020nuscenes}, a large-scale dataset for autonomous driving that includes 1.4 million camera images and 90,000 LiDAR sweeps across 1,000 scenes. Each point cloud keyframe in the nuScenes dataset is accompanied by six calibrated surround images. During the pre-training phase, we utilize the unlabeled RGB images and point clouds from 600 scenes to update the backbones in our pre-training model, following the same setting as SLidR~\citep{sautier2022slidr}. For fine-tuning across the three 3D perception tasks, we evaluate the performance of the pre-trained 3D backbone under different labeling percentages on the various datasets.

 \vspace{0.5em}
\noindent\textbf{Pre-training Details.} 
We pre-train the model for 50 epochs with an initial learning rate of 0.001, which is adjusted using a one-cycle learning rate scheduler~\citep{smith2017cyclical} and the Adam optimizer. The 2D backbone is a Vision Transformer (ViT) pretrained with DINOv2~\citep{oquab2023dinov2}. The pre-training tasks are performed on four RTX 3090 GPUs, with a batch size of sixteen. During pretraining, each image input is randomly masked, with 50\% of its content masked at each step. For occupancy estimation, we randomly select two thousand query points per scene. For fine-tuning, we adhere to the same data splits, augmentation strategies, and evaluation protocols as those used in prior works~\citep{mahmoud2023stslidr,chen2024building} on the nuScenes and SemanticKITTI datasets, and apply a similar procedure on other datasets.

\setlength{\tabcolsep}{6pt}
\begin{table*}[htp]
	\centering
	\vspace{-0.2cm}
	\caption{
		Comparison of various pre-training methods for semantic segmentation on the nuScenes and SemanticKITTI datasets, evaluated using both fine-tuning and linear probing (LP). The table reports the mean Intersection over Union (mIoU) scores on the validation set for different proportions of available annotations (1\%, 5\%, 10\%, 25\%, and 100\%) from both datasets.
	}
	\label{table:sem_ns_sk}
	%	\scalebox{0.95}{
		\begin{tabular}{l|c|cccccc|c}
			\Xhline{2\arrayrulewidth}
			\multirow{2}{*}{Method} & \multirow{2}{*}{Present at} &  \multicolumn{6}{c|}{nuScenes} & SemanticKITTI    \\
			& & LP & 1\%   & 5\%   & 10\%  & 25\%  & 100\%  & 1\% \\
			\hline
			Random     & -          & 8.1   & 30.30 & 47.84 & 56.15 & 65.48 & 74.66 & 39.50 \\
			PointContrast~\citep{xie2020pointcontrast}   & ECCV'20  & 21.90 & 32.50 & -     & -     & -     & -    & 41.10  \\
			DepthContrast~\citep{zhang2021depthcontrast} & ICCV'21  & 22.10 & 31.70 & -     & -     & -     & -   & 41.50   \\
			PPKT~\citep{liu2021ppkt}    & Arxiv'21 & 35.90 & 37.80 & 53.74 & 60.25 & 67.14 & 74.52 & 44.00 \\
			SLidR~\citep{sautier2022slidr} & CVPR'22 & 38.80 & 38.30 & 52.49 & 59.84 & 66.91 & 74.79 & 44.60 \\
			ST-SLidR~\citep{mahmoud2023stslidr} & CVPR'23 & 40.48 & 40.75 & 54.69 & 60.75 & 67.70 & 75.14 & 44.72 \\
                % HVDistill~\cite{zhang2024hvdistill}         & IJCV'24 & 39.50 & 42.70 & \underline{56.60} & 62.90 & \underline{69.30} & \textbf{76.60} \\
                Seal~\citep{liu2024seal}     & NeurIPS'23 & 44.95 & 45.84 & 55.64 & 62.97 & 68.41 & 75.60 & 46.63 \\
                CSC~\citep{chen2024building} & CVPR'24    & 46.00 & 47.00 & 57.00 & 63.30 & 68.60 & 75.70 & 47.20 \\
                SuperFlow~\citep{xu20254d}   & \major{ECCV}'24    & \underline{48.01} & \underline{49.95} & \underline{60.72} & \underline{65.09} & \underline{70.01} & \textbf{77.19} & \underline{49.07} \\
                \hline
			%		OLIVINE                    & NeurIPS'24 & \textbf{50.09} & \textbf{50.58} & \textbf{60.19} & \textbf{65.01} & \textbf{70.13} & \textbf{76.54} & \textbf{49.38} \\
			\rowcolor[rgb]{0.902,0.902,0.902} Ours & - & \textbf{51.23} & \textbf{51.76} & \textbf{61.08} & \textbf{65.69} & \textbf{70.72} & \underline{76.34} & \textbf{49.86} \\
			\Xhline{2\arrayrulewidth}
		\end{tabular}
		%	}
	%	\vspace{-0.3cm}
\end{table*}
\setlength{\tabcolsep}{6pt}

\setlength{\tabcolsep}{4pt}
\begin{table*}[htp]
	\centering
	\caption{Evaluation of various pretraining methods, initially trained on the nuScenes dataset, and fine-tuned on multiple downstream point cloud datasets. The table presents the mean Intersection over Union (mIoU) scores at different ratios of available annotations.}
	\label{table:sem_more_datasets}
	\scalebox{1}{
		\begin{tabular}{c|cc|cc|cc|cc|cc|cc} 
			\Xhline{2\arrayrulewidth}
			\multirow{2}{*}{Method} & \multicolumn{2}{c|}{ScribbleKITTI} & \multicolumn{2}{c|}{RELLIS-3D} & \multicolumn{2}{c|}{SemanticPOSS} & \multicolumn{2}{c|}{SemanticSTF} & \multicolumn{2}{c|}{SynLiDAR} & \multicolumn{2}{c}{DAPS-3D}  \\
			& 1\%   & 10\%   & 1\%   & 10\% & 50\%  & 100\%    & 50\%  & 100\%  & 1\%   & 10\%     & 50\%  & 100\%   \\ 
			\hline
			Random     & 23.81 & 47.60 & 38.46 & 53.60     & 46.26 & 54.12& 48.03 & 48.15       & 19.89 & 44.74    & 74.32 & 79.38   \\
			PPKT~\citep{liu2021ppkt} & 36.50 & 51.67 & 49.71 & 54.33     & 50.18 & 56.00& 50.92 & 54.69       & 37.57 & 46.48    & 78.90 & 84.00   \\
			SLidR~\citep{sautier2022slidr} & 39.60 & 50.45 & 49.75 & 54.57     & 51.56 & 55.36& 52.01 & 54.35       & 42.05 & 47.84    & 81.00 & 85.40   \\
			Seal~\citep{liu2024seal} & 40.64 & 52.77 & 51.09 & 55.03     & 53.26 & 56.89& 53.46 & 55.36       & 43.58 & 49.26    & 81.88 & 85.90   \\
                SuperFlow~\citep{xu20254d} & \underline{42.70} & \underline{54.00} & \underline{52.83} & \underline{55.71} & \underline{54.41} & \underline{57.33} & \underline{54.72} & \underline{56.57} & \underline{44.85} & \underline{51.38} & \underline{82.43} & \underline{86.21} \\
			\hline
			%			Ours       & \textbf{44.91} & \textbf{54.96} & \textbf{53.47} & \textbf{58.21}     & \textbf{55.70} & \textbf{58.51}& \textbf{56.65} & \textbf{60.42}       & \textbf{46.34} & \textbf{52.78}    & \textbf{83.63} & \textbf{86.84}   \\ % OLIVNE
			\rowcolor[rgb]{0.902,0.902,0.902} Ours & \textbf{45.29} & \textbf{55.36} & \textbf{54.87} & \textbf{56.40} &  \textbf{55.97} & \textbf{58.63} & \textbf{57.32} & \textbf{60.71} & \textbf{46.95} & \textbf{53.58} & \textbf{84.46} & \textbf{87.29} \\
			\Xhline{2\arrayrulewidth}
		\end{tabular}
	}
\end{table*}
\setlength{\tabcolsep}{6pt}

%\subsection{Comparison with State-of-the-Art Methods}\label{sec:exp_comparison}
\subsection{Transfer on 3D Semantic Segmentation}\label{sec:exp_semantic_segmentation}
In this section, we compare CMCR with several state-of-the-art 3D self-supervised learning methods across different benchmark datasets. Following the approach outlined in SLidR~\citep{sautier2022slidr}, we fine-tune the pre-trained 3D backbone using subsets of point cloud data with varying percentages of labeled annotations: 1\%, 5\%, 10\%, 25\%, and 100\% for nuScenes, and 1\% for SemanticKITTI. Additionally, we perform a linear evaluation using 100\% of the annotations, where only a linear classification head is trained while the rest of the 3D backbone layers are frozen. This helps assess the generalizability of the representations learned through self-supervised learning without task-specific fine-tuning. The evaluation is based on the mean Intersection over Union (mIoU) metric to compare the performance of different methods.

\noindent\textbf{Quantitative Results}. Under the linear probing setting, our method achieves the highest mIoU of 51.23\% on nuScenes, significantly outperforming the previous state-of-the-art method, CSC~\citep{chen2024building}, which records a mIoU of 46.00\% (see Table~\ref{table:sem_ns_sk}). This highlights the superior representation quality of our pre-trained features without task-specific fine-tuning. 
For fine-tuning on nuScenes, our method consistently excels, particularly in low-data regimes, achieving a mIoU of 51.76\% with just 1\% of annotations, significantly higher than CSC (47.00\%) and Seal~\citep{liu2024seal} (45.84\%). This trend persists across other proportions, with our method attaining 61.08\% mIoU at 5\% data and 65.69\% at 10\%, consistently outperforming all baselines. At 100\% annotation, our method achieves a mIoU of 76.34\%, surpassing CSC (75.70\%) and Seal (75.60\%). On SemanticKITTI, with only 1\% annotation, our method achieves 49.86\% mIoU, outperforming CSC (47.20\%) and Seal (46.63\%). These results demonstrate the robustness and versatility of our approach across diverse datasets and annotation settings.

As shown in Table~\ref{table:sem_more_datasets}, our method consistently achieves state-of-the-art performance across six point cloud datasets. At 1\% annotation, it achieves a mIoU of 45.29\% on ScribbleKITTI, outperforming Seal (40.64\%). For 10\% annotation on RELLIS-3D, it reaches 56.40\%, surpassing Seal’s 55.03\%. Even in fully supervised settings, such as 100\% annotation on DAPS-3D, our method achieves the highest mIoU of 87.29\%, compared to Seal’s 85.90\%. These results demonstrate the robustness and versatility of our approach across diverse datasets and annotation levels.

\begin{table}
	\centering
	\caption{
		\major{Performance of our pre-training method using the WaffleIron (WI-768) backbone. "LP" refers to linear probing with a frozen backbone. Results are reported as mIoU on the nuScenes dataset under different proportions of labeled data.}
	}
	\label{table:wi_backbone}	
	\begin{tabular}{c|cccc}
		\toprule
		Method          & LP        & 1\%   & 10\%  & 100\%  \\
		\midrule
		No pre-training & -         & 35.41 & 61.55 & 78.06  \\
		Ours            & \bf 65.82  & \textbf{53.64} & \textbf{71.89} & \textbf{78.93} \\
		\bottomrule
	\end{tabular}
\end{table}

\major{Furthermore, Table~\ref{table:wi_backbone} presents the results of applying our self-supervised method to the state-of-the-art WaffleIron (WI-768) backbone~\citep{puy2023using}. Under full supervision, WI-768 achieves a strong mIoU of 78.06\% on the nuScenes dataset. With our self-supervised pretraining, we achieve 71.89\% mIoU using only 10\% of the labeled data—substantially narrowing the gap to the fully supervised performance. Notably, our method attains 65.82\% mIoU through linear probing alone, which is only 12.24 points below the fully supervised result. These findings demonstrate that our approach—leveraging 2D image features without requiring manual 3D annotations—can significantly bridge the gap to fully supervised 3D representation quality.}

\begin{figure*}[htp]
	\centering
	\includegraphics[width=\textwidth]{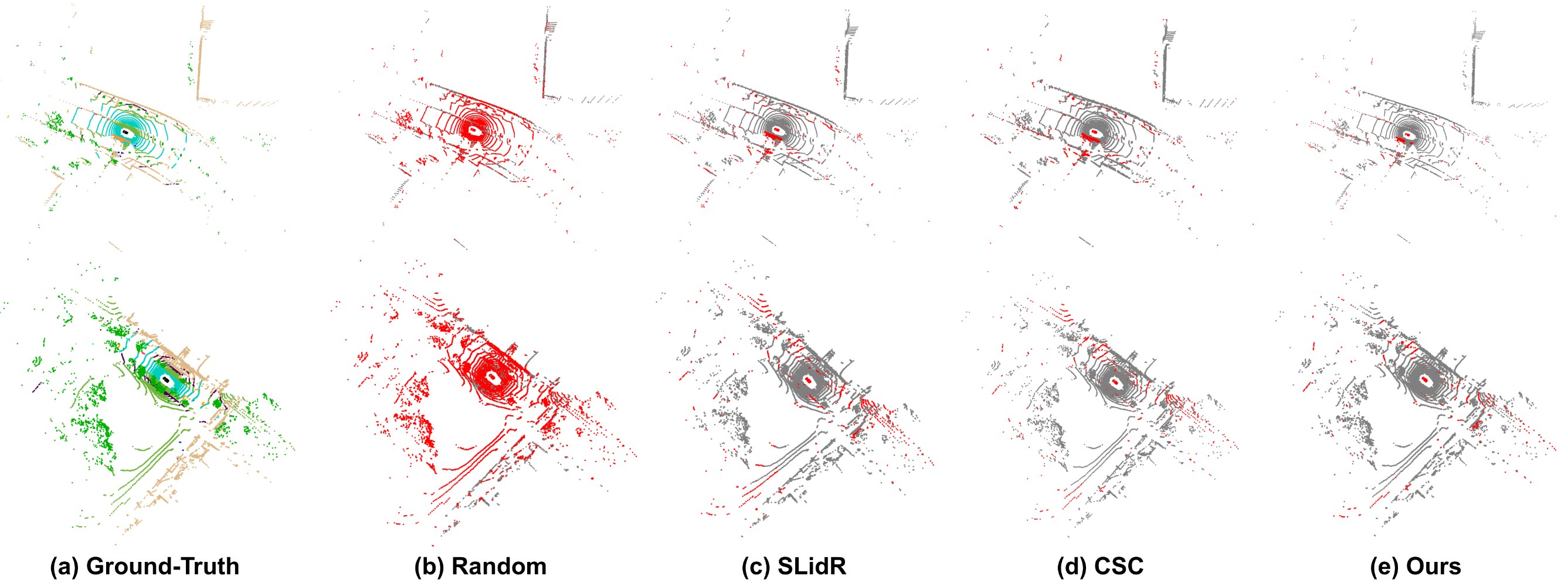} 
	\caption{
		The visual results of different point cloud pretraining methods, where the models were pre-trained on nuScenes and fine-tuned using only 1\% of annotated data. Correctly predicted areas are highlighted in gray, while incorrect predictions are marked in red to highlight the differences.
	}
	\label{fig:vis_results_nuscenes}
\end{figure*}

\noindent\textbf{Qualitative Results}. Figure~\ref{fig:vis_results_nuscenes} presents the visual comparison of various point cloud pre-training methods, all fine-tuned on nuScenes with just 1\% annotated data. The regions in gray represent areas where the model successfully predicted the semantic labels, while the red regions highlight incorrect predictions. This visualization demonstrates how our method outperforms other pretraining strategies in terms of accuracy and robustness, even with very limited labeled data. The improved segmentation quality, especially in complex urban environments, reflects the superior feature representations learned during pretraining.

\setlength{\tabcolsep}{4pt}
\begin{table}
	\centering
	\setlength{\extrarowheight}{0pt}
	\addtolength{\extrarowheight}{\aboverulesep}
	\addtolength{\extrarowheight}{\belowrulesep}
	\setlength{\aboverulesep}{0pt}
	\setlength{\belowrulesep}{0pt}
	\caption{
		Performance results (mAP and NDS) for fine-tuning pre-trained models on object detection tasks using two architectures (CenterPoint and SECOND) with varying amounts of labeled data (5\%, 10\%, and 20\%) on the nuScenes dataset.
	}
	\label{table:ann_eff_obj_det}
	\begin{tabular}{l|cc|cc|cc} 
		\Xhline{2\arrayrulewidth}
		\multirow{3}{*}{Method}                & \multicolumn{6}{c}{nuScenes}                                                                                                                                                                  \\ 
		\cmidrule{2-7}
		& \multicolumn{2}{c|}{5\%} & \multicolumn{2}{c|}{10\%} & \multicolumn{2}{c}{20\%}  \\
		& mAP   & NDS & mAP  & NDS  & mAP  & NDS     \\ 
		\hline
		\multicolumn{7}{l}{\textit{VoxelNet + CenterPoint}}                                                                                                                                                                                    \\
		No Pre-training   & 38.0  & 44.3 & 46.9  & 55.5  & 50.2  & 59.7   \\
		Point Con.     & 39.8  & 45.1 & 47.7  & 56.0  & -     & -      \\
		GCC-3D & 41.1  & 46.8 & 48.4  & 56.7  & -     & -      \\
		SLidR  & 43.3  & 52.4 & 47.5  & 56.8  & 50.4  & 59.9   \\
		TriCC  & 44.6  & \underline{54.4} & 48.9  & 58.1  & 50.9  & 60.3   \\
		CSC    & \underline{45.3}  & 54.2  & \underline{49.3} & \underline{58.3}  & \underline{51.9}  & \underline{61.3}   \\ 
		\rowcolor[rgb]{0.902,0.902,0.902} 
		Ours   & \textbf{46.6}  & \textbf{55.2}  & \textbf{50.2} & \textbf{59.1}  & \textbf{52.7} & \textbf{62.0}   \\ 
		\hline
		\multicolumn{7}{l}{\textit{VoxelNet + SECOND}}   \\
		No Pre-training   & 35.8  & 45.9 & 39.0  & 51.2  & 43.1  & 55.7   \\
		SLidR  & 36.6  & 48.1 & 39.8  & 52.1  & 44.2  & 56.3   \\
		TriCC  & 37.8  & \underline{50.0} & 41.4  & 53.5  & 45.5  & 57.7   \\
		CSC & \underline{38.2} & 49.4 & \underline{42.5} & \underline{54.8} & \underline{45.6} & \underline{58.1}  \\
		\rowcolor[rgb]{0.902,0.902,0.902}
		%		Ours & \textbf{38.7} & 49.4 & \textbf{42.5} & \textbf{54.8} & \textbf{45.6} & \textbf{58.1}  \\
		Ours & \textbf{39.4} & 	\textbf{50.7} & 	\textbf{43.5} & 	\textbf{55.7} & 	\textbf{46.5} & 	\textbf{59.1} \\ 
		\Xhline{2\arrayrulewidth}
	\end{tabular}
	\vspace{-0.4cm}
\end{table}
\setlength{\tabcolsep}{6pt}

\vspace{-0.3cm}
\subsection{Transfer on 3D Object Detection}\label{sec:exp_detection}
To assess the effectiveness of our pre-trained lidar representation for 3D object detection task, we conduct experiments on the nuScenes dataset. Specifically, we fine-tune the pre-trained backbone using varying percentages of labeled data: 5\%, 10\%, and 20\%. For evaluation, we incorporate the pre-trained model into two state-of-the-art object detection architectures, CenterPoint~\citep{yin2021center} and SECOND~\citep{yan2018second}, and report the mean average precision (mAP) and nuScenes detection score (NDS). NDS is a composite metric that integrates mAP with additional factors, providing a comprehensive assessment of model performance.

As shown in Table \ref{table:ann_eff_obj_det}, we present the results of our method alongside other existing approaches. We evaluate the performance of different methods on both CenterPoint and SECOND detection models, with varying amounts of labeled data. For the CenterPoint-based model, our method outperforms all alternatives across all labeled data conditions. With only 5\% labeled data, our approach achieves a notable improvement in both mAP (46.6\%) and NDS (55.2\%) compared to the second-best method, CSC, which achieved 45.3\% mAP and 54.2\% NDS. This performance gap increases further as more labeled data is available, with our method achieving the highest scores across all annotation settings (10\% and 20\%), with 50.2\% mAP and 59.1\% NDS at 10\% annotations, and 52.7\% mAP and 62.0\% NDS at 20\% annotations.

Similarly, for the SECOND-based model, our approach continues to outperform others, achieving the highest mAP and NDS scores across all annotation levels. At 5\% labeled data, our method achieves 39.4\% mAP and 50.7\% NDS, surpassing the second-best model (CSC) by 1.2\% in mAP and 1.3\% in NDS. As the amount of labeled data increases, the performance improvement remains consistent, with our method achieving 43.5\% mAP and 55.7\% NDS at 10\% annotations, and 46.5\% mAP and 59.1\% NDS at 20\% annotations.

\vspace{-0.3cm}
\subsection{Transfer on 3D Panoptic Segmentation}\label{sec:exp_panoptic}
In this part, we evaluate the effectiveness of various pre-training strategies for panoptic segmentation, a task that requires both semantic and instance recognition capabilities. 

We fine-tune the pre-trained models on the nuScenes dataset with 1\% and 5\% labeled data, comparing our approach to existing methods. 
Our method, which leverages the power of the pre-trained 3D representations, is compared against other approaches including random initialization and prior pre-training techniques such as SLidR~\citep{sautier2022slidr} and more advanced methods like CSC~\citep{chen2024building}.
In evaluating our pre-trained model for 3D panoptic segmentation, following CSC~\citep{chen2024building}, we use Panoptic-PolarNet~\citep{zhou2021panoptic} with Cylinder3D~\citep{zhu2021cylindrical} as the backbone, following the current state-of-the-art supervised methods in 3D panoptic segmentation.

\revise{As shown in Table \ref{table:ann_eff_pan_seg}, our method consistently outperforms the others across all evaluation metrics: Panoptic Quality (PQ), Segmentation Quality (SQ), and Recognition Quality (RQ). At the 1\% label setting, our approach achieves the highest scores in all three metrics, with a PQ of 20.7, SQ of 80.5, and RQ of 26.1, surpassing the second-best model (SuperFlow) by 0.9\% PQ, 2.2\% SQ, and 1.0\% RQ. This performance improvement remains consistent with 5\% labeled data, where our method achieves a PQ of 24.0\%, SQ of 78.5\%, and RQ of 29.3\%, outpacing SuperFlow again by 0.6\% PQ, 0.9\% SQ, and 0.5\% RQ.}

\setlength{\tabcolsep}{4pt}
\begin{table}[t]
	\centering
	%	\caption{\todo{Results (PQ, SQ, and RQ) when fine-tuning the pre-trained models to panoptic segmentation with 1\% and 5\% labels on nuScenes.}} 	
	\caption{\revise{Panoptic segmentation results (PQ, SQ, and RQ) after fine-tuning pre-trained models on the nuScenes dataset with 1\% and 5\% labeled data.}}
	\label{table:ann_eff_pan_seg}
	\begin{tabular}{l | c c c | c c c }
		\toprule
		\multirow{3}{*}{Method} 
		& 
		\multicolumn{6}{c}{nuScenes} 
		\\ 
		\cmidrule{2-7} 
		& \multicolumn{3}{c}{1\%}  \vline & \multicolumn{3}{c}{5\%}   
		\\
		{} & PQ   & SQ & RQ & PQ   & SQ & RQ
		\\ 
		\midrule
		No Pre-training & 15.3 & 62.6 & 20.4 & 20.9 & 73.4 & 26.5
		\\
		SLidR + SLIC & 16.3 & 65.7 & 21.4 & 21.6 & 73.5 & 27.1
		\\
		SLidR + DINOV2 & 17.6 & 70.7 & 22.7 & 22.3 & 75.1 & 27.8
		\\
		CSC & 19.3	& 74.5	& 24.6	& 23.1	& 76.9 &	28.5 
		\\
		\revise{SuperFlow} & \underline{19.8} & \underline{78.3} & \underline{25.1} & \underline{23.4} & \underline{77.6} & \underline{28.8}
		\\
		\rowcolor{gray!20}
		Ours & \bf20.7 & \bf80.5 & \bf26.1 & \bf24.0 & \bf 78.5 & \bf 29.3 \\
		\bottomrule
	\end{tabular}
\end{table}
\setlength{\tabcolsep}{6pt}

The overall improvements suggest that our method not only enhances the network's ability to segment and recognize objects more effectively but also achieves a balanced boost across both semantic and instance recognition tasks. The gains in SQ are particularly notable, which indicates that our approach significantly improves the model’s ability to accurately classify semantic categories. The improvements in RQ highlight that our method also helps with better distinguishing individual instances within those categories.

\subsection{Ablation Study}\label{sec:ablation_study}

\setlength{\tabcolsep}{2.4pt}
\begin{table}[t]
	\centering
	\caption{Ablation study of each component pre-trained and fine-tuned on nuScenes. \textbf{PP}: Basic pixel-point contrastive learning. \textbf{Rec.}: Image reconstruction and occupancy estimation tasks. \textbf{Codebook}: The multi-modal unified codebook. \textbf{Geo.}: Geometry-enhanced masked image modeling.}
	\label{table:ablation}
	\begin{tabular}{ c| c c c c c| c c c |c}
		\toprule
		\multirow{2}{*}{Exp.} & \multirow{2}{*}{PP} & \multirow{2}{*}{Rec.} & \multirow{2}{*}{Codebook} & \multirow{2}{*}{Geo.} & \multirow{2}{*}{$\mathcal{L}_{\text{kl}}$} & \multicolumn{3}{c|}{nuScene} & S.K.  \\
		%		\cline{6-9}
		& & & & & & LP & 1\% & 5\% & 1\%\\
		\midrule
		(1) & \checkmark & & &  & & 38.5 & 40.4 & 52.4 & 43.1 \\
		(2) & \checkmark & \checkmark & & &  & 43.4 & 43.8 & 55.3 & 45.9 \\
		(3) & \checkmark & \checkmark & \checkmark & &  & 45.6 & 44.5 & 56.6 & 46.5 \\
		(4) & \checkmark & \checkmark & \checkmark & \checkmark & & 46.3 & 45.7 & 57.5 & 47.1 \\
		(5) & \checkmark & \checkmark & \checkmark & \checkmark & \checkmark & \bf 51.2 & \bf 51.7 & \bf 61.1 & \bf 49.8 \\
		\bottomrule
	\end{tabular}
\end{table}
\setlength{\tabcolsep}{6pt}

\noindent\textbf{Effect of Key Components.}The results of our ablation study, presented in Table~\ref{table:ablation}, highlight the impact of each key component on the performance of the model pre-trained and fine-tuned on the nuScenes dataset. In Experiment (1), using only the basic pixel-point contrastive learning (PP) leads to relatively low performance, with the model achieving an LP score of 38.5 and 1\% and 5\% fine-tuning scores of 40.4 and 52.4, respectively. Adding the image reconstruction and occupancy estimation tasks (Rec.) in Experiment (2) provides a noticeable performance boost across all metrics, with the LP score improving to 43.4 and the fine-tuning scores reaching 43.8 for 1\% and 55.3 for 5\%, demonstrating the positive contribution of these auxiliary tasks.
When the multi-modal unified codebook is introduced in Experiment (3), the performance improves further, with the LP score increasing to 45.6 and the fine-tuning results reaching 44.5 for 1\% and 56.6 for 5\%. This suggests that the integration of a shared codebook for both image and point cloud modalities improves cross-modal alignment, helping the model better capture common representations. The addition of geometry-enhanced masked image modeling in Experiment (4) brings additional gains, particularly for fine-tuning with 1\% and 5\% labeled data, where the scores reach 45.7 and 57.5, respectively. This indicates that geometry-based enhancements further improve the model's ability to leverage spatial context for multi-modal fusion. Finally, Experiment (5) demonstrates the full model achieves the highest performance across all metrics, with a significant jump in the LP score to 51.2 and 1\% and 5\% fine-tuning scores of 51.7 and 61.1, respectively. Each component contributes to the overall performance, with the complete model achieving the best results for both pre-training and fine-tuning tasks.

\begin{figure}[htp]
	\centering
	\includegraphics[width=0.5\textwidth]{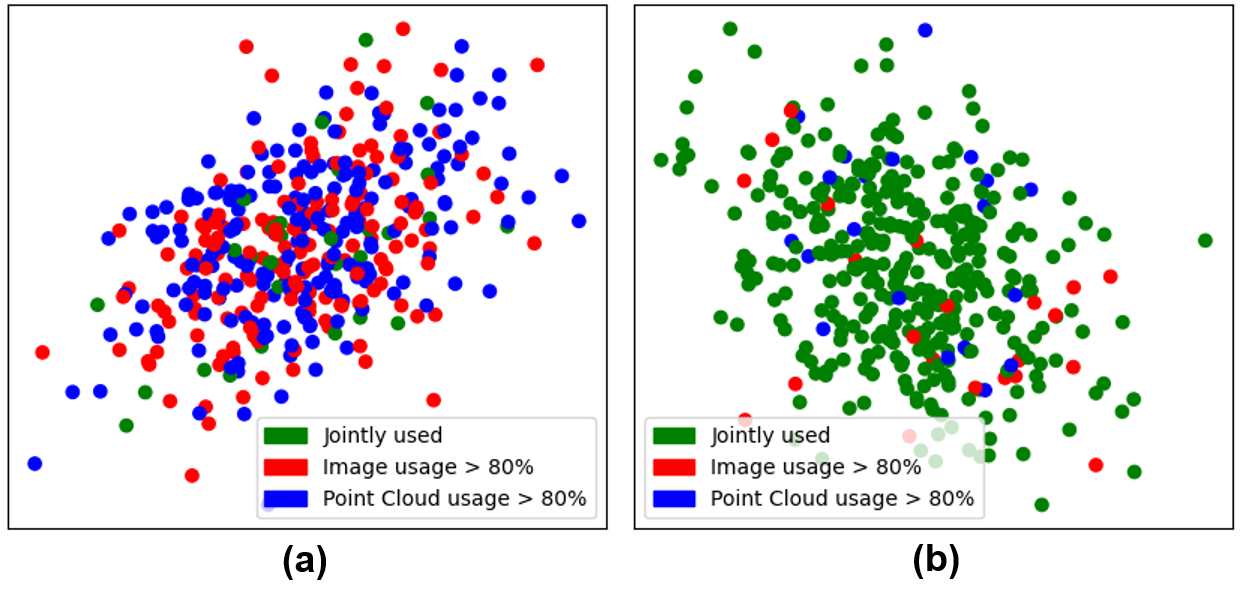} 
	\caption{
		T-SNE visualization of the codebook with and without the proposed Multi-modal Unified Codebook.
		On the left, without the multi-modal unified codebook, red points represent codewords primarily used by the image modality, while blue points represent those primarily used by the point cloud modality. Green points indicate codewords that are jointly used by both modalities. On the right, with the proposed multi-modal unified codebook, codewords are more evenly distributed, with a larger proportion of jointly used codewords (green), highlighting the improved cross-modal alignment.
	}
	\label{fig:codebook_tsne}
\end{figure}

\noindent\textbf{Effect of Multi-modal Unified Codebook.} 
After incorporating the proposed Multi-modal Unified Codebook, we observe a significant improvement in the distribution of codewords across modalities. As shown in the T-SNE visualization in Fig.~\ref{fig:codebook_tsne}, without the unified codebook, the codewords are largely segregated, with red points representing codewords predominantly used by the image modality and blue points used mainly by the point cloud modality. Only a few green points indicate codewords that are jointly shared by both modalities. In contrast, with the Multi-modal Unified Codebook, the codewords are more evenly distributed, and a larger proportion of codewords (green points) are now jointly utilized by both modalities. This demonstrates the enhanced cross-modal alignment facilitated by the unified codebook, which enables the model to more effectively capture shared semantics between the 2D and 3D modalities, leading to a better fusion of information across both inputs.

\setlength{\tabcolsep}{10pt}
\begin{table}[t]
	\centering
	\caption{The effect of different codebook sizes.}
	\label{table:ablation_codebook}
	\begin{tabular}{c|ccc} 
		\toprule
		\multirow{2}{*}{Codebook Size} & \multicolumn{3}{c}{nuScene}  \\
		& LP  &  1\% &  5\%     \\ 
		\midrule
		128  & 45.3  & 44.0  & 54.2 \\ 
		256  & 48.1  & 47.2  & 58.3 \\ 
		512  & \textbf{51.2}  & \textbf{51.7}  & \textbf{61.1} \\ 
		1024 & 49.5  & 50.3  & 59.8 \\ 
		2048 & 47.6  & 48.5  & 58.1 \\ 
		\bottomrule
	\end{tabular}
\end{table}
\setlength{\tabcolsep}{6pt}

\vspace{0.5em}
\noindent\textbf{Effect of Codebook Size.} The size of the codebook plays a crucial role in balancing the aggregation of modality-specific features, which directly impacts the performance of downstream tasks. As shown in Table \ref{table:ablation_codebook}, smaller codebook sizes (e.g., 128 and 256) result in lower performance, likely due to the insufficient capacity to capture the complex relationships between modalities. On the other hand, larger sizes (e.g., 1024 and 2048) do not consistently yield significant improvements and can lead to overfitting or misalignment of modality-specific features. The optimal performance is achieved with a codebook size of 512, where the model exhibits the best trade-off between representation capacity and cross-modal generalization. This suggests that a moderate codebook size is critical for maximizing the effectiveness of modality fusion, ensuring that both shared and specific features are captured accurately without introducing redundancy or misalignment.

\begin{figure}[b]
	\centering
	\includegraphics[width=0.45\textwidth]{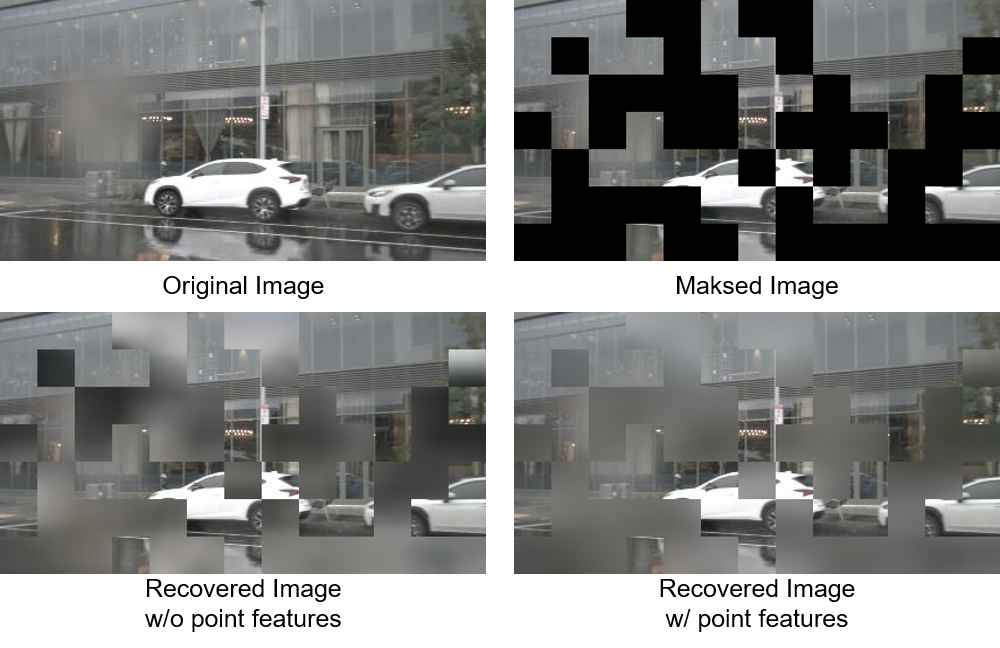} 
	\caption{
		\major{Impact of point cloud features on recovered image quality.}
%		Effectiveness of Point Cloud Features in Image Recovery
	}
	\label{fig:recovered_images}
\end{figure}

\vspace{0.5em}
\noindent\textbf{\major{Effect of Geometry Enhanced Masked Image Modeling.}} 
%We have included visualizations to demonstrate the contribution of point cloud features to the image reconstruction process. Figure~\ref{fig:recovered_images} compares the recovered images with and without the use of point cloud features. The results show that incorporating point cloud features helps the reconstructed image better resemble the original image overall.
\major{We have included visualizations to illustrate the impact of point cloud features on the image reconstruction process. Figure~\ref{fig:recovered_images} compares the recovered images with and without the use of point cloud features. The results show that incorporating point cloud features significantly improves the overall resemblance of the reconstructed image to the original.}

\vspace{0.5em}
\noindent\textbf{\major{Effect of 2D Backbone Settings.}} 
\major{Thank you for your insightful comment. To evaluate the effect of masked image modeling (MIM), we conducted an ablation study with two settings: (1) freezing the 2D backbone, and (2) making the 2D backbone trainable with MIM task enabled. As shown in Table~\ref{table:ablation_2d_backbone}, allowing the 2D backbone to be updated and jointly training it with MIM leads to consistent improvements across all settings (e.g., +2.7 mIoU in linear probing, +2.2 mIoU with 5\% labeled data on nuScenes). It is important to emphasize that MIM is not merely an auxiliary supervision task. As discussed in Section~\ref{sec:analysis}, conventional contrastive learning methods primarily focus on learning modality-shared features while neglecting modality-specific cues. The inclusion of MIM in our framework encourages the image branch to capture modality-specific features, which are complementary to the modality-shared ones. This results in richer, more comprehensive representations that improve downstream 3D task performance.}

\setlength{\tabcolsep}{6pt}
\begin{table}[t]
	\centering
	\caption{
		\major{Ablation study on the effect of 2D backbone settings, comparing frozen vs. trainable backbones and the impact of masked image modeling (MIM) on representation quality.}
	}
	\label{table:ablation_2d_backbone}
	\begin{tabular}{cc|ccc|c} 
		\toprule
		\multirow{2}{*}{Frozen} & \multirow{2}{*}{MIM} & \multicolumn{3}{c|}{nuScene} & S.K.  \\
		&      & LP   & 1\%  & 5\%   & 1\%   \\ 
		\midrule 
		\checkmark	&  $\times$    & 48.5 &	48.8 & 58.9 & 48.0 \\
		$\times$	&   \checkmark & \bf 51.2 & \bf 51.7 & \bf 61.1 & \bf 49.8 \\
		\bottomrule
	\end{tabular}
\end{table}
\setlength{\tabcolsep}{6pt}

\section{Conclusion}\label{sec:conclusion}
In this paper, we have presented a novel framework for learning more comprehensive 3D representations. By addressing the limitations of existing methods that focus primarily on modality-shared features, we introduced a new approach that also captures modality-specific features through masked image modeling and occupancy estimation tasks. Our key contribution, the multi-modal unified codebook, enables the learning of a shared embedding space across different modalities, facilitating better cross-modal alignment and representation learning. Moreover, the geometry-enhanced masked image modeling further enhances 3D representation learning by incorporating spatial structure information.
Through extensive experiments, we demonstrated that our approach significantly improves upon traditional methods and outperforms existing image-to-LiDAR contrastive distillation methods in downstream tasks. Future work could explore additional task-specific adaptations and further optimizations to improve the performance of multi-modal 3D learning.

\section*{Acknowledgements}
% This work was supported in part by the NSFC Excellent Young Scientists Fund 62422118, in part by the Hong Kong Research Grants Council under Grants 11219324 and 11219422, and in part by Shanghai Pujiang Program 25PJA042. This work was also supported by Shanghai Technical Service Center of Science and Engineering Computing, Shanghai University.
This work was supported in part by the National Natural Science Foundation of China Excellent Young Scientists Fund (Grant 62422118); in part by the Hong Kong Research Grants Council (Grants 11219324 and 11219422); and in part by the Shanghai Pujiang Program (Grant 25PJA042).
% It was also supported by Shanghai Technical Service Center of Science and Engineering Computing, Shanghai University.

\section*{Data Availability}
This work does not propose a new dataset. All the datasets we used are publicly available from the papers cited in Appendix~\textcolor{red}{A}.

\section*{Conflict of Interest}
The authors affirm that there are no commercial or associative relationships that could be perceived as a conflict of interest related to the submitted work.

\bibliographystyle{spbasic}
\bibliography{bib}

\begin{thebibliography}{69}
\providecommand{\natexlab}[1]{#1}
\providecommand{\url}[1]{{#1}}
\providecommand{\urlprefix}{URL }
\expandafter\ifx\csname urlstyle\endcsname\relax
  \providecommand{\doi}[1]{DOI~\discretionary{}{}{}#1}\else
  \providecommand{\doi}{DOI~\discretionary{}{}{}\begingroup
  \urlstyle{rm}\Url}\fi
\providecommand{\eprint}[2][]{\url{#2}}

\bibitem[{Behley et~al.(2019)Behley, Garbade, Milioto, Quenzel, Behnke,
  Stachniss, and Gall}]{behley2019semantickitti}
Behley J, Garbade M, Milioto A, Quenzel J, Behnke S, Stachniss C, Gall J (2019)
  Semantickitti: A dataset for semantic scene understanding of lidar sequences.
  In: Proceedings of the IEEE/CVF International Conference on Computer Vision,
  pp 9297--9307

\bibitem[{Berman et~al.(2018)Berman, Triki, and Blaschko}]{berman2018lovasz}
Berman M, Triki AR, Blaschko MB (2018) The lov{\'a}sz-softmax loss: A tractable
  surrogate for the optimization of the intersection-over-union measure in
  neural networks. In: Proceedings of the IEEE/CVF Conference on Computer
  Vision and Pattern Recognition, pp 4413--4421

\bibitem[{Boulch et~al.(2023)Boulch, Sautier, Michele, Puy, and
  Marlet}]{boulch2023ALSO}
Boulch A, Sautier C, Michele B, Puy G, Marlet R (2023) Also: Automotive lidar
  self-supervision by occupancy estimation. In: Proceedings of the IEEE/CVF
  Conference on Computer Vision and Pattern Recognition, pp 13455--13465

\bibitem[{Caesar et~al.(2020)Caesar, Bankiti, Lang, Vora, Liong, Xu, Krishnan,
  Pan, Baldan, and Beijbom}]{caesar2020nuscenes}
Caesar H, Bankiti V, Lang AH, Vora S, Liong VE, Xu Q, Krishnan A, Pan Y, Baldan
  G, Beijbom O (2020) nuscenes: A multimodal dataset for autonomous driving.
  In: Proceedings of the IEEE/CVF Conference on Computer Vision and Pattern
  Recognition, pp 11621--11631

\bibitem[{Caron et~al.(2021)Caron, Touvron, Misra, J{\'e}gou, Mairal,
  Bojanowski, and Joulin}]{caron2021emerging}
Caron M, Touvron H, Misra I, J{\'e}gou H, Mairal J, Bojanowski P, Joulin A
  (2021) Emerging properties in self-supervised vision transformers. In:
  Proceedings of the IEEE/CVF International Conference on Computer Vision, pp
  9650--9660

\bibitem[{Chen et~al.(2022{\natexlab{a}})Chen, Chen, Zhang, and
  Tao}]{chen2022sasa}
Chen C, Chen Z, Zhang J, Tao D (2022{\natexlab{a}}) Sasa: Semantics-augmented
  set abstraction for point-based 3d object detection. In: Proceedings of the
  AAAI Conference on Artificial Intelligence, vol~36, pp 221--229

\bibitem[{Chen et~al.(2024)Chen, Zhang, Qu, Zhang, Tan, and
  Xie}]{chen2024building}
Chen H, Zhang Z, Qu Y, Zhang R, Tan X, Xie Y (2024) Building a strong
  pre-training baseline for universal 3d large-scale perception. In:
  Proceedings of the IEEE/CVF Conference on Computer Vision and Pattern
  Recognition, pp 19925--19935

\bibitem[{Chen et~al.(2020)Chen, Fan, Girshick, and He}]{chen2020improved}
Chen X, Fan H, Girshick R, He K (2020) Improved baselines with momentum
  contrastive learning. arXiv preprint arXiv:200304297

\bibitem[{Chen et~al.(2022{\natexlab{b}})Chen, Nießner, and
  Dai}]{chen20224dcontrast}
Chen Y, Nießner M, Dai A (2022{\natexlab{b}}) 4dcontrast: Contrastive learning
  with dynamic correspondences for 3d scene understanding. In: European
  Conference on Computer Vision, pp 543--560

\bibitem[{Chen et~al.(2023)Chen, Yuan, Tian, Geng, Li, Zhou, Metaxas, and
  Yang}]{chen2023revisiting}
Chen Y, Yuan J, Tian Y, Geng S, Li X, Zhou D, Metaxas DN, Yang H (2023)
  Revisiting multimodal representation in contrastive learning: from patch and
  token embeddings to finite discrete tokens. In: Proceedings of the IEEE/CVF
  Conference on Computer Vision and Pattern Recognition, pp 15095--15104

\bibitem[{Chibane et~al.(2022)Chibane, Engelmann, Anh~Tran, and
  Pons-Moll}]{chibane2022box2mask}
Chibane J, Engelmann F, Anh~Tran T, Pons-Moll G (2022) Box2mask: Weakly
  supervised 3d semantic instance segmentation using bounding boxes. In:
  European Conference on Computer Vision, pp 681--699

\bibitem[{Choe et~al.(2022)Choe, Park, Rameau, Park, and
  Kweon}]{choe2022pointmixer}
Choe J, Park C, Rameau F, Park J, Kweon IS (2022) Pointmixer: Mlp-mixer for
  point cloud understanding. In: European Conference on Computer Vision,
  Springer, pp 620--640

\bibitem[{Choy et~al.(2019)Choy, Gwak, and Savarese}]{choy20194d}
Choy C, Gwak J, Savarese S (2019) 4d spatio-temporal convnets: Minkowski
  convolutional neural networks. In: Proceedings of the IEEE/CVF Conference on
  Computer Vision and Pattern Recognition, pp 3075--3084

\bibitem[{Fadadu et~al.(2022)Fadadu, Pandey, Hegde, Shi, Chou, Djuric, and
  Vallespi-Gonzalez}]{fadadu2022multi}
Fadadu S, Pandey S, Hegde D, Shi Y, Chou FC, Djuric N, Vallespi-Gonzalez C
  (2022) Multi-view fusion of sensor data for improved perception and
  prediction in autonomous driving. In: Proceedings of the IEEE/CVF Winter
  Conference on Applications of Computer Vision, pp 2349--2357

\bibitem[{Feder and Merhav(1994)}]{feder1994relations}
Feder M, Merhav N (1994) Relations between entropy and error probability. IEEE
  Transactions on Information theory 40(1):259--266

\bibitem[{He et~al.(2022)He, Chen, Xie, Li, Doll{\'a}r, and
  Girshick}]{he2022masked}
He K, Chen X, Xie S, Li Y, Doll{\'a}r P, Girshick R (2022) Masked autoencoders
  are scalable vision learners. In: Proceedings of the IEEE/CVF Conference on
  Computer Vision and Pattern Recognition, pp 16000--16009

\bibitem[{Hess et~al.(2023)Hess, Jaxing, Svensson, Hagerman, Petersson, and
  Svensson}]{hess2023masked}
Hess G, Jaxing J, Svensson E, Hagerman D, Petersson C, Svensson L (2023) Masked
  autoencoder for self-supervised pre-training on lidar point clouds. In:
  Proceedings of the IEEE/CVF Winter Conference on Applications of Computer
  Vision, pp 350--359

\bibitem[{Ho et~al.(2023)Ho, Tai, Lin, Yang, and Tsai}]{ho2024diffusion}
Ho CJ, Tai CH, Lin YY, Yang MH, Tsai YH (2023) Diffusion-ss3d: Diffusion model
  for semi-supervised 3d object detection. Advances in Neural Information
  Processing Systems 36:49100--49112

\bibitem[{Huang et~al.(2021)Huang, Xie, Zhu, and Zhu}]{huang2021STRL}
Huang S, Xie Y, Zhu SC, Zhu Y (2021) Spatio-temporal self-supervised
  representation learning for 3d point clouds. In: Proceedings of the IEEE/CVF
  International Conference on Computer Vision, pp 6535--6545

\bibitem[{Jiang et~al.(2021)Jiang, Osteen, Wigness, and
  Saripallig}]{jiang2021rellis3D}
Jiang P, Osteen P, Wigness M, Saripallig S (2021) Rellis-3d dataset: Data,
  benchmarks and analysis. In: IEEE International Conference on Robotics and
  Automation, pp 1110--1116

\bibitem[{Klokov et~al.(2023)Klokov, Pak, Khorin, Yudin, Kochiev, Luchinskiy,
  and Bezuglyj}]{klokov2023daps3D}
Klokov AA, Pak DU, Khorin A, Yudin DA, Kochiev L, Luchinskiy VD, Bezuglyj VD
  (2023) Daps3d: Domain adaptive projective segmentation of 3d lidar point
  clouds. IEEE Access 11:79341--79356

\bibitem[{Kong et~al.(2023{\natexlab{a}})Kong, Liu, Chen, Ma, Zhu, Li, Hou,
  Qiao, and Liu}]{kong2023rethinking}
Kong L, Liu Y, Chen R, Ma Y, Zhu X, Li Y, Hou Y, Qiao Y, Liu Z
  (2023{\natexlab{a}}) Rethinking range view representation for lidar
  segmentation. In: Proceedings of the IEEE/CVF International Conference on
  Computer Vision, pp 228--240

\bibitem[{Kong et~al.(2023{\natexlab{b}})Kong, Ren, Pan, and
  Liu}]{kong2023lasermix}
Kong L, Ren J, Pan L, Liu Z (2023{\natexlab{b}}) Lasermix for semi-supervised
  lidar semantic segmentation. In: Proceedings of the IEEE/CVF Conference on
  Computer Vision and Pattern Recognition, pp 21705--21715

\bibitem[{Liang et~al.(2023)Liang, Deng, Ma, Zou, Morency, and
  Salakhutdinov}]{liang2024factorized}
Liang PP, Deng Z, Ma MQ, Zou JY, Morency LP, Salakhutdinov R (2023) Factorized
  contrastive learning: Going beyond multi-view redundancy. Advances in Neural
  Information Processing Systems 36:32971--32998

\bibitem[{Liao et~al.(2024)Liao, Li, and Ye}]{liao2024vlm2scene}
Liao G, Li J, Ye X (2024) Vlm2scene: Self-supervised image-text-lidar learning
  with foundation models for autonomous driving scene understanding. In:
  Proceedings of the AAAI Conference on Artificial Intelligence, vol~38, pp
  3351--3359

\bibitem[{Liu et~al.(2022{\natexlab{a}})Liu, Jin, Lai, Rouditchenko, Oliva, and
  Glass}]{LiuJLROG22}
Liu AH, Jin S, Lai C, Rouditchenko A, Oliva A, Glass JR (2022{\natexlab{a}})
  Cross-modal discrete representation learning. In: Proceedings of the 60th
  Annual Meeting of the Association for Computational Linguistics, pp
  3013--3035

\bibitem[{Liu et~al.(2023{\natexlab{a}})Liu, Zhan, Zhang, Xu, Yu, El~Saddik,
  Theobalt, Xing, and Lu}]{liu2023weakly}
Liu K, Zhan F, Zhang J, Xu M, Yu Y, El~Saddik A, Theobalt C, Xing E, Lu S
  (2023{\natexlab{a}}) Weakly supervised 3d open-vocabulary segmentation.
  Advances in Neural Information Processing Systems 36:53433--53456

\bibitem[{Liu et~al.(2022{\natexlab{b}})Liu, Zhou, Qi, Gong, Su, and
  Anguelov}]{liu2022less}
Liu M, Zhou Y, Qi CR, Gong B, Su H, Anguelov D (2022{\natexlab{b}}) Less:
  Label-efficient semantic segmentation for lidar point clouds. In: European
  Conference on Computer Vision, pp 70--89

\bibitem[{Liu et~al.(2023{\natexlab{b}})Liu, Kong, Cen, Chen, Zhang, Pan, Chen,
  and Liu}]{liu2024seal}
Liu Y, Kong L, Cen J, Chen R, Zhang W, Pan L, Chen K, Liu Z
  (2023{\natexlab{b}}) Segment any point cloud sequences by distilling vision
  foundation models. Advances in Neural Information Processing Systems
  36:37193--37229

\bibitem[{Liu et~al.(2021)Liu, Huang, Chiang, Su, Liu, Chen, Tseng, and
  Hsu}]{liu2021ppkt}
Liu YC, Huang YK, Chiang HY, Su HT, Liu ZY, Chen CT, Tseng CY, Hsu WH (2021)
  Learning from 2d: Contrastive pixel-to-point knowledge transfer for 3d
  pretraining. arXiv preprint arXiv:210404687

\bibitem[{Luo et~al.(2023)Luo, Chen, Wang, Yu, Huang, and
  Baktashmotlagh}]{luo2023exploring}
Luo Y, Chen Z, Wang Z, Yu X, Huang Z, Baktashmotlagh M (2023) Exploring active
  3d object detection from a generalization perspective. In: The Eleventh
  International Conference on Learning Representations, pp 1--13

\bibitem[{Mahmoud et~al.(2023)Mahmoud, Hu, Kuai, Harakeh, Paull, and
  Waslander}]{mahmoud2023stslidr}
Mahmoud A, Hu JS, Kuai T, Harakeh A, Paull L, Waslander SL (2023)
  Self-supervised image-to-point distillation via semantically tolerant
  contrastive loss. In: Proceedings of the IEEE/CVF Conference on Computer
  Vision and Pattern Recognition, pp 7102--7110

\bibitem[{Nunes et~al.(2022)Nunes, Marcuzzi, Chen, Behley, and
  Stachniss}]{nunes2022segcontrast}
Nunes L, Marcuzzi R, Chen X, Behley J, Stachniss C (2022) Segcontrast: 3d point
  cloud feature representation learning through self-supervised segment
  discrimination. IEEE Robotics and Automation Letters 7(2):2116--2123

\bibitem[{Nunes et~al.(2023)Nunes, Wiesmann, Marcuzzi, Chen, Behley, and
  Stachniss}]{nunes2023TARL}
Nunes L, Wiesmann L, Marcuzzi R, Chen X, Behley J, Stachniss C (2023) Temporal
  consistent 3d lidar representation learning for semantic perception in
  autonomous driving. In: Proceedings of the IEEE/CVF Conference on Computer
  Vision and Pattern Recognition, pp 5217--5228

\bibitem[{Oord et~al.(2018)Oord, Li, and Vinyals}]{oord2018representation}
Oord Avd, Li Y, Vinyals O (2018) Representation learning with contrastive
  predictive coding. arXiv preprint arXiv:180703748

\bibitem[{Oquab et~al.(2024)Oquab, Darcet, Moutakanni, Vo, Szafraniec,
  Khalidov, Fernandez, Haziza, Massa, El-Nouby et~al.}]{oquab2023dinov2}
Oquab M, Darcet T, Moutakanni T, Vo H, Szafraniec M, Khalidov V, Fernandez P,
  Haziza D, Massa F, El-Nouby A, et~al. (2024) Dinov2: Learning robust visual
  features without supervision. Trans Mach Learn Res 2024:1--28

\bibitem[{Pan et~al.(2020)Pan, Gao, Mei, Geng, Li, and
  Zhao}]{pan2020semanticPOSS}
Pan Y, Gao B, Mei J, Geng S, Li C, Zhao H (2020) Semanticposs: A point cloud
  dataset with large quantity of dynamic instances. In: IEEE Intelligent
  Vehicles Symposium, pp 687--693

\bibitem[{Pang et~al.(2023)Pang, Xia, and Lu}]{pang2023unsupervised}
Pang B, Xia H, Lu C (2023) Unsupervised 3d point cloud representation learning
  by triangle constrained contrast for autonomous driving. In: Proceedings of
  the IEEE/CVF Conference on Computer Vision and Pattern Recognition, pp
  5229--5239

\bibitem[{Pang et~al.(2022)Pang, Wang, Tay, Liu, Tian, and
  Yuan}]{pang2022masked}
Pang Y, Wang W, Tay FE, Liu W, Tian Y, Yuan L (2022) Masked autoencoders for
  point cloud self-supervised learning. In: European Conference on Computer
  Vision, pp 604--621

\bibitem[{Peng et~al.(2022)Peng, Dong, Bao, Ye, and Wei}]{peng2022beit}
Peng Z, Dong L, Bao H, Ye Q, Wei F (2022) Beit v2: Masked image modeling with
  vector-quantized visual tokenizers. arXiv preprint arXiv:220806366

\bibitem[{Poursaeed et~al.(2020)Poursaeed, Jiang, Qiao, Xu, and
  Kim}]{poursaeed2020self}
Poursaeed O, Jiang T, Qiao H, Xu N, Kim VG (2020) Self-supervised learning of
  point clouds via orientation estimation. In: International Conference on 3D
  Vision, pp 1018--1028

\bibitem[{Puy et~al.(2023)Puy, Boulch, and Marlet}]{puy2023using}
Puy G, Boulch A, Marlet R (2023) Using a waffle iron for automotive point cloud
  semantic segmentation. In: Proceedings of the IEEE/CVF International
  Conference on Computer Vision, pp 3379--3389

\bibitem[{Puy et~al.(2024)Puy, Gidaris, Boulch, Sim\'eoni, Sautier, P\'erez,
  Bursuc, and Marlet}]{puy24scalr}
Puy G, Gidaris S, Boulch A, Sim\'eoni O, Sautier C, P\'erez P, Bursuc A, Marlet
  R (2024) Three pillars improving vision foundation model distillation for
  lidar. In: Proceedings of the IEEE/CVF Conference on Computer Vision and
  Pattern Recognition, pp 21519--21529

\bibitem[{Ronneberger et~al.(2015)Ronneberger, Fischer, and
  Brox}]{ronneberger2015unet}
Ronneberger O, Fischer P, Brox T (2015) U-net: Convolutional networks for
  biomedical image segmentation. In: Medical Image Computing and Computer
  Assisted Intervention, Springer, pp 234--241

\bibitem[{Sauder and Sievers(2019)}]{sauder2019self}
Sauder J, Sievers B (2019) Self-supervised deep learning on point clouds by
  reconstructing space. In: Advances in Neural Information Processing Systems,
  vol~32, pp 12942--12952

\bibitem[{Sautier et~al.(2022)Sautier, Puy, Gidaris, Boulch, Bursuc, and
  Marlet}]{sautier2022slidr}
Sautier C, Puy G, Gidaris S, Boulch A, Bursuc A, Marlet R (2022) Image-to-lidar
  self-supervised distillation for autonomous driving data. In: Proceedings of
  the IEEE/CVF Conference on Computer Vision and Pattern Recognition, pp
  9891--9901

\bibitem[{Smith(2017)}]{smith2017cyclical}
Smith LN (2017) Cyclical learning rates for training neural networks. In:
  Proceedings of the IEEE/CVF Winter Conference on Applications of Computer
  Vision, IEEE, pp 464--472

\bibitem[{Sridharan and Kakade(2008)}]{sridharan2008information}
Sridharan K, Kakade SM (2008) An information theoretic framework for multi-view
  learning. In: Annual Conference on Computational Learning Theory, 114, pp
  403--414

\bibitem[{Team et~al.(2020)}]{od2020openpcdet}
Team O, et~al. (2020) Openpcdet: An open-source toolbox for 3d object detection
  from point clouds

\bibitem[{Tian et~al.(2022)Tian, Chu, Wang, Wei, and Shen}]{tian2022fully}
Tian Z, Chu X, Wang X, Wei X, Shen C (2022) Fully convolutional one-stage 3d
  object detection on lidar range images. Advances in Neural Information
  Processing Systems 35:34899--34911

\bibitem[{Unal et~al.(2022)Unal, Dai, and Gool}]{unal2022scribbleKITTI}
Unal O, Dai D, Gool LV (2022) Scribble-supervised lidar semantic segmentation.
  In: Proceedings of the IEEE/CVF Conference on Computer Vision and Pattern
  Recognition, pp 2697--2707

\bibitem[{Van Den~Oord et~al.(2017)Van Den~Oord, Vinyals
  et~al.}]{van2017neural}
Van Den~Oord A, Vinyals O, et~al. (2017) Neural discrete representation
  learning. In: Advances in Neural Information Processing Systems, pp
  6306--6315

\bibitem[{Xia et~al.(2023)Xia, Huang, Zhu, and Zhao}]{xia2024achieving}
Xia Y, Huang H, Zhu J, Zhao Z (2023) Achieving cross modal generalization with
  multimodal unified representation. Advances in Neural Information Processing
  Systems 36:63529--63541

\bibitem[{Xiao et~al.(2022)Xiao, Huang, Guan, Zhan, and Lu}]{xiao2022synLiDAR}
Xiao A, Huang J, Guan D, Zhan F, Lu S (2022) Transfer learning from synthetic
  to real lidar point cloud for semantic segmentation. In: AAAI Conference on
  Artificial Intelligence, pp 2795--2803

\bibitem[{Xiao et~al.(2023)Xiao, Huang, Xuan, Ren, Liu, Guan, Saddik, Lu, and
  Xing}]{xiao2023semanticSTF}
Xiao A, Huang J, Xuan W, Ren R, Liu K, Guan D, Saddik AE, Lu S, Xing E (2023)
  3d semantic segmentation in the wild: Learning generalized models for
  adverse-condition point clouds. In: Proceedings of the IEEE/CVF Conference on
  Computer Vision and Pattern Recognition, pp 9382--9392

\bibitem[{Xie et~al.(2023)Xie, Li, Guo, Liu, and Cheng}]{xie2023annotator}
Xie B, Li S, Guo Q, Liu C, Cheng X (2023) Annotator: A generic active learning
  baseline for lidar semantic segmentation. Advances in Neural Information
  Processing Systems 36:48444--48458

\bibitem[{Xie et~al.(2020)Xie, Gu, Guo, Qi, Guibas, and
  Litany}]{xie2020pointcontrast}
Xie S, Gu J, Guo D, Qi CR, Guibas L, Litany O (2020) Pointcontrast:
  Unsupervised pre-training for 3d point cloud understanding. In: European
  Conference on Computer Vision, pp 574--591

\bibitem[{Xu et~al.(2013)Xu, Tao, and Xu}]{xu2013survey}
Xu C, Tao D, Xu C (2013) A survey on multi-view learning. arXiv preprint
  arXiv:13045634

\bibitem[{Xu et~al.(2021)Xu, Zhang, Dou, Zhu, Sun, and Pu}]{xu2021rpvnet}
Xu J, Zhang R, Dou J, Zhu Y, Sun J, Pu S (2021) Rpvnet: A deep and efficient
  range-point-voxel fusion network for lidar point cloud segmentation. In:
  Proceedings of the IEEE/CVF International Conference on Computer Vision, pp
  16024--16033

\bibitem[{Xu et~al.(2025)Xu, Kong, Shuai, Zhang, Pan, Chen, Liu, and
  Liu}]{xu20254d}
Xu X, Kong L, Shuai H, Zhang W, Pan L, Chen K, Liu Z, Liu Q (2025) 4d
  contrastive superflows are dense 3d representation learners. In: European
  Conference on Computer Vision, pp 58--80

\bibitem[{Yan et~al.(2018)Yan, Mao, and Li}]{yan2018second}
Yan Y, Mao Y, Li B (2018) Second: Sparsely embedded convolutional detection.
  Sensors 18(10):3337

\bibitem[{Yin et~al.(2022)Yin, Zhou, Zhang, Fang, Xu, Shen, and
  Wang}]{yin2022proposalcontrast}
Yin J, Zhou D, Zhang L, Fang J, Xu CZ, Shen J, Wang W (2022) Proposalcontrast:
  Unsupervised pre-training for lidar-based 3d object detection. In: European
  Conference on Computer Vision, pp 17--33

\bibitem[{Yin et~al.(2021)Yin, Zhou, and Krahenbuhl}]{yin2021center}
Yin T, Zhou X, Krahenbuhl P (2021) Center-based 3d object detection and
  tracking. In: Proceedings of the IEEE/CVF Conference on Computer Vision and
  Pattern Recognition, pp 11784--11793

\bibitem[{Zhang et~al.(2024)Zhang, Deng, Bai, Li, Ouyang, and
  Zhang}]{zhang2024hvdistill}
Zhang S, Deng J, Bai L, Li H, Ouyang W, Zhang Y (2024) Hvdistill: Transferring
  knowledge from images to point clouds via unsupervised hybrid-view
  distillation. International Journal of Computer Vision 132(7):2585--2599

\bibitem[{Zhang and Hou(2024)}]{zhang2024fine}
Zhang Y, Hou J (2024) Fine-grained image-to-lidar contrastive distillation with
  visual foundation models. Advances in Neural Information Processing Systems
  37:128396--128429

\bibitem[{Zhang et~al.(2021)Zhang, Girdhar, Joulin, and
  Misra}]{zhang2021depthcontrast}
Zhang Z, Girdhar R, Joulin A, Misra I (2021) Self-supervised pretraining of 3d
  features on any point-cloud. In: Proceedings of the IEEE/CVF International
  Conference on Computer Vision, pp 10252--10263

\bibitem[{Zhou and Tuzel(2018)}]{zhou2018voxelnet}
Zhou Y, Tuzel O (2018) Voxelnet: End-to-end learning for point cloud based 3d
  object detection. In: Proceedings of the IEEE/CVF Conference on Computer
  Vision and Pattern Recognition, pp 4490--4499

\bibitem[{Zhou et~al.(2021)Zhou, Zhang, and Foroosh}]{zhou2021panoptic}
Zhou Z, Zhang Y, Foroosh H (2021) Panoptic-polarnet: Proposal-free lidar point
  cloud panoptic segmentation. In: Proceedings of the IEEE/CVF Conference on
  Computer Vision and Pattern Recognition, pp 13194--13203

\bibitem[{Zhu et~al.(2021)Zhu, Zhou, Wang, Hong, Ma, Li, Li, and
  Lin}]{zhu2021cylindrical}
Zhu X, Zhou H, Wang T, Hong F, Ma Y, Li W, Li H, Lin D (2021) Cylindrical and
  asymmetrical 3d convolution networks for lidar segmentation. In: Proceedings
  of the IEEE/CVF Conference on Computer Vision and Pattern Recognition, pp
  9939--9948

\end{thebibliography}

\section*{Appendix}
\section*{A. Datasets}\label{appendix:datasets}
\noindent\textbf{NuScenes Dataset.} The NuScenes dataset, gathered from driving recordings in Boston and Singapore, is equipped with a 32-beam LiDAR and other sensing technologies~\citep{caesar2020nuscenes}. It represents a comprehensive autonomous vehicle sensor array, featuring a 32-beam LiDAR, six cameras, and radar systems to capture a full 360-degree view of the environment. The dataset contains 850 driving clips, with 700 scenes for training and 150 for validation. Each scene spans 20 seconds, with annotations provided every 0.5 seconds. Extensive object category annotations are included, covering vehicles, pedestrians, bicycles, and road barriers, each represented by 3D bounding boxes and augmented with attributes like visibility, activity, and pose. The extended NuScenes-lidarseg dataset enhances the original NuScenes with semantic and panoptic segmentation annotations~\citep{caesar2020nuscenes}. This version includes semantic labels for 32 distinct categories, with each point in keyframes precisely annotated. We utilize the 700 training scenes with segmentation annotations to fine-tune semantic segmentation models, evaluating them on the 150 validation scenes.

\vspace{0.5em}
\noindent\textbf{SemanticKITTI Dataset.} The SemanticKITTI dataset offers paired RGB images and point cloud data from KITTI’s urban environments~\citep{behley2019semantickitti}, specifically curated for semantic segmentation tasks. The dataset is collected using vehicle-mounted sensors, comprising over 200,000 images and corresponding point clouds across 21 distinct sequences. Images are captured at 1241x376 resolution, and each point cloud contains roughly 40,000 3D points. Both modalities are aligned to maintain consistent relative transformations. The dataset is split into 10 training sequences and a single validation sequence (the eighth sequence).

\newcommand*\rotext{\multicolumn{1}{R{60}{0.7em}}}
\setlength{\tabcolsep}{4pt}
\begin{table*}[b]
    \centering
    \caption{
        Per-class IoU results on the nuScenes dataset, fine-tuned using only 1\% of the labeled data. The table displays the Intersection over Union (IoU) scores for each category, with the highest and second-highest values highlighted in bold and underlined, respectively.
    }
    \label{table:per_class_result_ns}
        \scalebox{1}{
	\begin{tabular}{l|cccccccccccccccc|c}
		\Xhline{2\arrayrulewidth}
		Method & \rotext{barrier} & \rotext{bicycle} & \rotext{bus} & \rotext{car} & \rotext{const. veh.} & \rotext{motor} & \rotext{pedestrian} & \rotext{traffic cone} & \rotext{trailer} & \rotext{truck} & \rotext{driv. surf.} & \rotext{other flat} & \rotext{sidewalk} & \rotext{terrain} & \rotext{manmade} & \rotext{vegetation} & \rotext{\textbf{mIoU}} \\
		\hline
		Random & 0.0 & 0.0 & 8.1 & 65.0 & 0.1 & 6.6 & 21.0 & 9.0 & 9.3 & 25.8 & 89.5 & 14.8 & 41.7 & 48.7 & 72.4 & 73.3 & 30.3 \\
		PointContrast & 0.0& 1.0 & 5.6 & 67.4 & 0.0 & 3.3 & 31.6 & 5.6 & 12.1 & 30.8 & 91.7 & 21.9 & 48.4 & 50.8 & 75.0 & 74.6 & 32.5 \\
		DepthContrast & 0.0& 0.6 & 6.5 & 64.7 & 0.2 & 5.1 & 29.0 & 9.5 & 12.1 & 29.9 & 90.3 & 17.8 & 44.4 & 49.5 & 73.5 & 74.0 & 31.7 \\
		PPKT & 0.0& 2.2 & 20.7 & 75.4 & 1.2 & 13.2 & 45.6 & 8.5 & 17.5 & 38.4 & 92.5 & 19.2 & 52.3 & 56.8 & 80.1 & 80.9 & 37.8 \\
		SLidR & 0.0& 1.8 & 15.4 & 73.1 & 1.9 & 19.9 & 47.2 & 17.1 & 14.5 & 34.5 & 92.0 & 27.1 & 53.6 & 61.0 & 79.8 & 82.3 & 38.3 \\
		ST-SLidR & 0.0& 2.7 & 16.0 & 74.5 & 3.2 & 25.4 & 50.9 & 20.0 & 17.7 & 40.2 & 92.0 & 30.7 & 54.2 & \underline{61.1} & 80.5 & \underline{82.9} & 40.8 \\
		Seal & 0.0 & \underline{9.4} & \underline{32.6} & \underline{77.5} & \underline{10.4} & \underline{28.0} & \underline{53.0} & \underline{25.0} & \textbf{30.9} & \underline{49.7} & \textbf{94.0} & \underline{33.7} & \textbf{60.1} & 59.6 & \textbf{83.9} & \textbf{83.4} & \underline{45.8} \\
		\hline
		Ours & \textbf{0.1} & \textbf{9.8} & \textbf{70.7} & \textbf{83.6} & \textbf{29.6} & \textbf{46.3} & \textbf{58.2} & \textbf{32.5} & \underline{19.6} & \textbf{52.1} & \underline{93.8} & \textbf{42.8} & \underline{59.6} & \textbf{64.7} & \underline{81.4} & 82.3 & \textbf{51.7}  \\
		\Xhline{2\arrayrulewidth}
	\end{tabular}
    }
\end{table*}

\renewcommand*\rotext{\multicolumn{1}{R{60}{1em}}}
\setlength{\tabcolsep}{3.4pt}
\begin{table*}[b]
	\centering
	\caption{
		Per-class IoU results on the SemanticKITTI dataset, fine-tuned using only 1\% of the labeled data. The table displays the Intersection over Union (IoU) scores for each category, with the highest and second-highest values highlighted in bold and underlined, respectively.
	}
	\label{table:per_class_result_sk}	
	\scalebox{0.95}{
		\begin{tabular}{c|ccccccccccccccccccc|c}
			\Xhline{2\arrayrulewidth}    
			Method   & \rotext{car} &\rotext{bicycle} &\rotext{motorcycle} &\rotext{truck} & \rotext{other-vehicle} &\rotext{person} &\rotext{bicyclist} &\rotext{motorcyclist} &\rotext{road} &\rotext{parking} & \rotext{sidewalk} & \rotext{other-ground}  & \rotext{building} & \rotext{fence} &\rotext{vegetation} &\rotext{trunk} & \rotext{terrain} &\rotext{pole}& \rotext{traffic-sign} &\rotext{\textbf{mIoU}}  \\
			\hline
			Random   & 91.2 & 0.0       & 9.4        & 8.0     & 10.7          & 21.2   & 0.0         & 0.0            & \underline{89.4} & 21.4   & 73.0  & \textbf{1.1}          & 85.3     & 41.1  & 84.9       & 50.1  & \underline{71.4}    & 55.4 & 37.6         & 39.5      \\
			PPKT     & 91.3 & 1.9     & 11.2       & \underline{23.1}  & 12.1          & 27.4   & \textbf{37.3}      & 0.0 & \textbf{91.3} & 27.0 & 74.6 & 0.3 & \underline{86.5} & 38.2  & \underline{85.3}       & 58.2  & \textbf{71.6} & 57.7 & 40.1 & 43.9      \\
			SLidR    & 92.2 & 3.0 & 17.0  & 22.4  & 14.3   & 36.0     & 22.1      & 0.0 & \textbf{91.3} & \underline{30.0} & 74.7 & 0.2   & \textbf{87.7}     & \underline{41.2}  & 85.0   & 58.5  & 70.4    & \underline{58.3} & 42.4         & 44.6      \\
			Seal & \underline{92.3} & \underline{14.9} & \underline{18.7} & 16.1 & \textbf{23.7} & \underline{43.0} & \underline{34.4} & 0.0 & \textbf{91.3} & 27.2 & \underline{75.3} & \underline{0.7} & 85.7 & 38.8 & 85.1 & \textbf{61.9} & 71.3 & 57.7 & \textbf{47.7} & \underline{46.6} \\
			\hline
%			OLIVINE     & 93.1 & 17.5 & 28.1 & 45.2 & 18.7     & 47.4     & 31.4     & 0.0        & 91.8     & 32.3     & 75.5     & 1.8      & 88.1 & 47.2 & 85.7 & 59.0 & 71.4 & 59.8 & 43.9 & 49.4 \\
			Ours & \textbf{93.5} & \textbf{19.0} & \textbf{22.7} & \textbf{41.4} & \underline{18.7} & \textbf{48.7} & 33.7 & 0.0 & \textbf{91.3} & \textbf{32.8} & \textbf{75.7} & 0.4 & \textbf{87.7} & \textbf{46.3} & \textbf{86.0} & \underline{60.0} & 71.3 & \textbf{60.8} & \underline{42.8} & \textbf{49.8} \\
			\Xhline{2\arrayrulewidth}    
		\end{tabular}
	}
\end{table*}
\setlength{\tabcolsep}{6pt}

%\vspace{0.5em}
\noindent\textbf{ScribbleKITTI Dataset.} ScribbleKITTI is derived from SemanticKITTI but features weak supervision, where only line scribbles (rather than fully labeled point clouds) are provided~\citep{unal2022scribbleKITTI}. The dataset retains the same 19,130 LiDAR scans captured with a Velodyne HDL-64E sensor, but semantic labels are provided for only 8.06\% of the points. This annotation method significantly reduces labeling time by approximately 90\%. The dataset is used to test the generalization of models pre-trained on fully annotated datasets with weak annotations. We follow the SLidR protocol to create different training splits, selecting one scan every 100 frames for 1\% labeled samples, with model performance evaluated on the official validation set.

\noindent\textbf{RELLIS-3D Dataset.} The RELLIS-3D dataset, collected in off-road environments on the Texas A\&M University campus, provides 13,556 annotated LiDAR scans~\citep{jiang2021rellis3D}. This dataset presents a challenging scenario with complex terrain and class imbalance, making it valuable for testing models in outdoor environments with varying topographies and object densities.

\noindent\textbf{SemanticPOSS Dataset.} The SemanticPOSS dataset focuses on dynamic objects and is captured on the Peking University campus~\citep{pan2020semanticPOSS}. It contains 2,988 LiDAR scans collected using a Hesai Pandora 40-channel LiDAR sensor. This dataset emphasizes moving objects and dense environments, making it suitable for evaluating model adaptability in dynamic scenes. In our experiments, sequences 00 and 01 provide half of the annotated training samples, while sequences 00-05 (excluding 02) are used for validation.

\noindent\textbf{SemanticSTF Dataset.} The SemanticSTF dataset includes 2,076 LiDAR scans captured under challenging weather conditions such as snow, fog, and rain, using a Velodyne HDL64 S3D sensor~\citep{xiao2023semanticSTF}. The dataset is split into training, validation, and test sets, with balanced weather conditions across all subsets. It is particularly useful for evaluating model robustness in extreme environmental conditions.

\noindent\textbf{SynLiDAR Dataset.} The SynLiDAR dataset comprises synthetic point clouds generated in virtual environments using Unreal Engine 4~\citep{xiao2022synLiDAR}. It consists of 13 sequences and 198,396 scans, providing a controlled setting for large-scale experimentation. The synthetic data closely mimics real-world scenarios, making it ideal for pre-training and testing models. For fine-tuning, we use a uniformly downsampled subset.

\noindent\textbf{DAPS-3D Dataset.} The DAPS-3D dataset includes both semi-synthetic and real-world data, with the DAPS-1 subset containing over 23,000 labeled LiDAR scans across 11 sequences~\citep{klokov2023daps3D}. Collected during an autonomous robot deployment in real-world environments, this dataset helps evaluate the transferability of models trained on synthetic data to real-world applications. We use sequence `38-18\_7\_72\_90` for training and validate on sequences `38-18\_7\_72\_90`, `42-48\_10\_78\_90`, and `44-18\_11\_15\_32` to assess model performance across both synthetic and real-world data.

\section*{B. Implementation Details}
\noindent\textbf{Network Architectures.}  
In our pre-training pipeline, the input images are resized to 416$\times$224. For the 3D semantic segmentation task, we utilize the Sparse Residual 3D U-Net 34 (SR-UNet34)~\citep{ronneberger2015unet}, following the methodology outlined in SLidR~\citep{sautier2022slidr}. The SR-UNet34 outputs a feature map with 256 channels, while the image branch produces a 64-dimensional feature vector. To match the dimensionality of these features, we employ a 3D convolutional layer in the projection head to reduce the point feature map to 64 channels. For input, the 3D point data is converted into voxels, with Cartesian coordinates spanning an X-Y range of [-51.2m, 51.2m] and a Z-range of [-5.0m, 3.0m]. Each voxel has dimensions of (0.1m, 0.1m, 0.1m).  
For 3D object detection, we adopt VoxelNet~\citep{zhou2018voxelnet}, with a maximum of 10 points per voxel and a limit of 60,000 voxels to process the point cloud input.

%\vspace{0.5em}
\noindent\textbf{Evaluation Protocol.} 
For the 3D semantic segmentation task, we build the network by incorporating a 3D convolutional layer as the segmentation head, which is appended to the pre-trained backbone. In line with prior studies~\citep{sautier2022slidr,mahmoud2023stslidr}, we fine-tune the network for 100 epochs using a batch size of 16 for nuScenes and 10 for other semantic segmentation datasets. The initial learning rates for the backbone and the segmentation head are set to 0.05 and 2.0, respectively. During fine-tuning, we explore different ratios of annotated data. Additionally, we assess the quality of the learned representation through a \textit{linear probing} protocol, where, unlike fine-tuning, only the newly added segmentation head is optimized while the weights of the backbone \( f_{\mathrm{3D}} \) are kept frozen, using the nuScenes dataset. In both protocols, the training objective is a weighted sum of the cross-entropy loss and the Lovász-Softmax loss~\citep{berman2018lovasz}. For the 3D object detection task, we adopt the default configuration from OpenPCDet~\citep{od2020openpcdet} and initialize the backbone with the pre-trained weights from our model. For the panoptic segmentation downstream task, the following hyperparameters were used in our setup: a batch size of 8, the Adam optimizer with an initial learning rate of 0.004, and a learning rate scheduler configured with the ``MultiStepLR" strategy. The learning rate milestones were set at epochs 30, 50, and 80, with a learning rate decay factor (\(\gamma\)) of 0.5. The model was trained for a total of 100 epochs.

\begin{figure}[htp]
	\centering
	\includegraphics[width=0.48\textwidth]{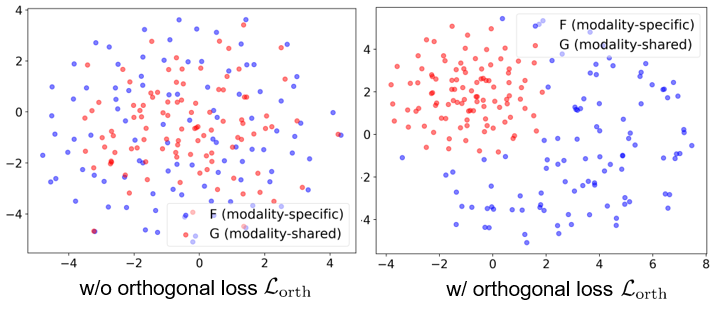} 
	\caption{
		\major{Visualization of constructed modality-shared and modality-specific features, without and with the orthogonal loss $\mathcal{L}_{\text{orth}}$.}
	}
	\label{fig:orth_comparison}
\end{figure}

\section*{C. Additional Experimental Results}
\noindent\textbf{Additional Quantitative Results.} Tables~\ref{table:per_class_result_ns} and~\ref{table:per_class_result_sk} present the per-class performance of various point cloud pretraining methods, including our approach and several baseline models, all fine-tuned with just 1\% of labeled data from the nuScenes-lidarseg and SemanticKITTI datasets. Our method demonstrates consistent superiority over other approaches, achieving the highest mean Intersection over Union (mIoU) scores in nearly all categories.

\major{As shown in Table~\ref{table:3d_occ_estimation}, incorporating image features slightly improves the 3D occupancy estimation performance (from 56.56 mIoU to 57.34 mIoU).  However, while image features can provide minor improvements in the occupancy estimation task, they do not significantly contribute to 3D representation learning. Therefore, we do not leverage image features to improve the 3D occupancy estimation task in the pipeline, as our primary focus is on enhancing 3D representation learning.}

\begin{table}[t]
	\centering
	\caption{\major{Comparison of 3D occupancy estimation performance without and with image features.}}
	\label{table:3d_occ_estimation}
	\begin{tabular}{c|c}
		\toprule
		Image Features & mIoU   \\
		\midrule		
		$\times$             & 56.56  \\
		\checkmark     & \textbf{57.34} \\
		\bottomrule
	\end{tabular}
\end{table}

%\vspace{0.5em}
\noindent\textbf{Additional Qualitative Results.}
\major{As shown in Figure~\ref{fig:orth_comparison}, the modality-specific and modality-shared features are clearly separated when the orthogonal loss $\mathcal{L}_{\text{orth}}$ is applied. But the features are mixed without the application of the orthogonal loss, showing some overlap and correlation between the features. The results demonstrate the effectiveness of the orthogonal loss in promoting disentangled representations.}

\begin{figure*}[htp]
	\centering
	\includegraphics[width=0.8\textwidth]{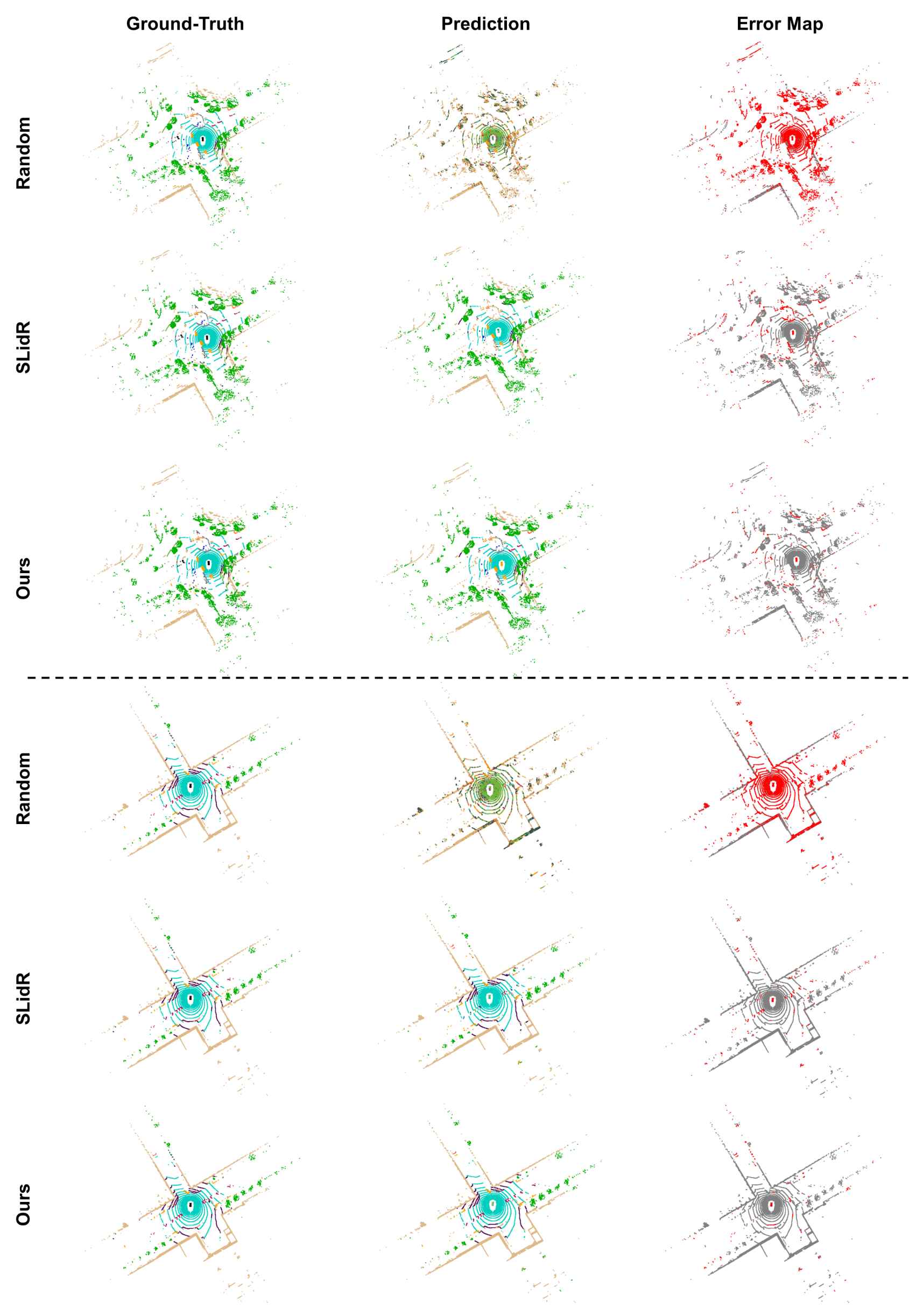} 
	\caption{
		Qualitative results of fine-tuning using 1\% of the nuScenes dataset with various pre-training approaches. The error maps on the right highlight incorrect predictions, marked in red. For optimal viewing, please refer to the color version and zoom in for greater detail.
	}
	\label{fig:vis_results_nuscenes_appendix}
\end{figure*}

\begin{figure*}[htp]
	\centering
	\includegraphics[width=0.9\textwidth]{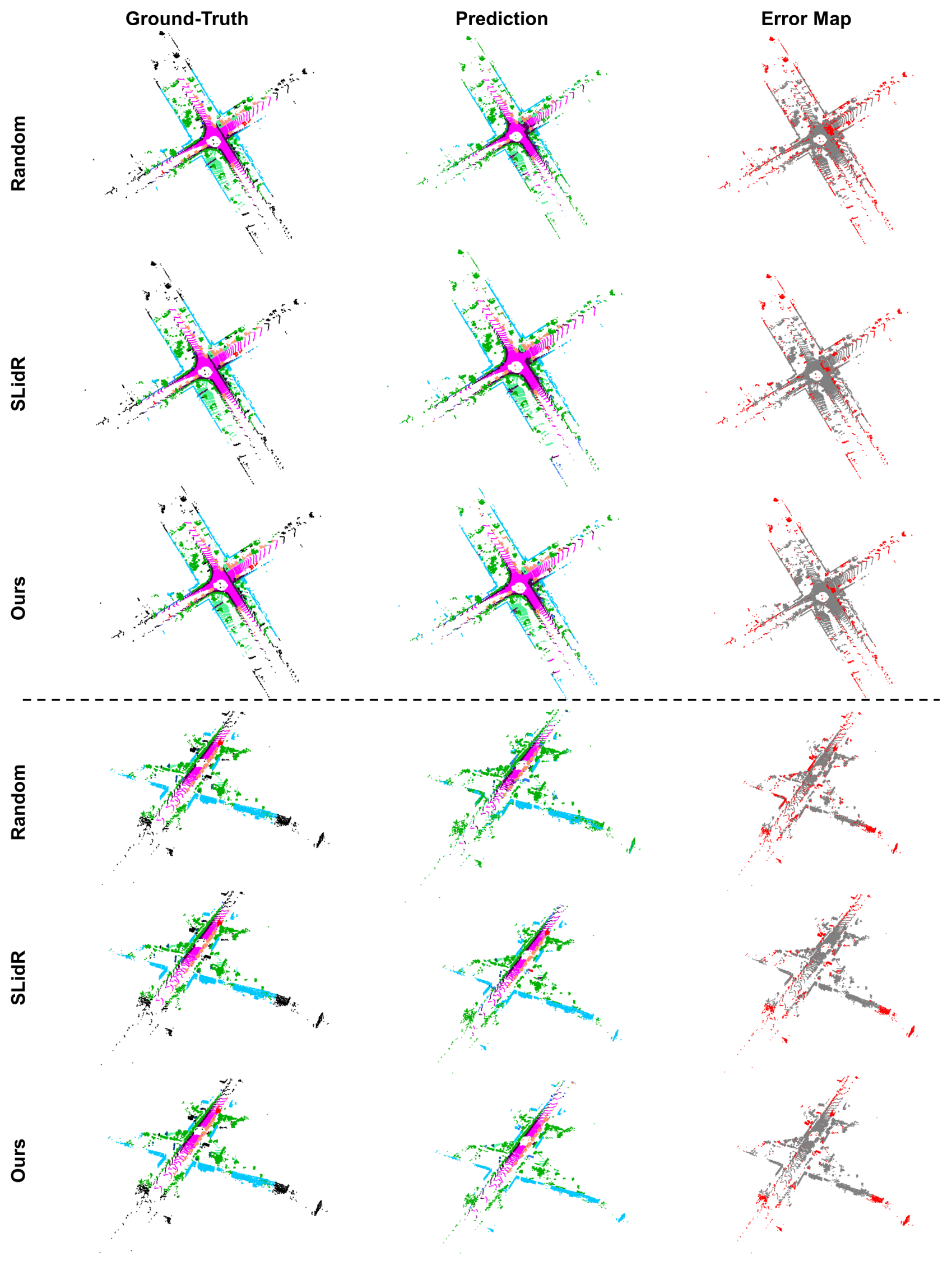} 
	\caption{
		Qualitative results of fine-tuning using 1\% of the SemanticKITTI dataset with various pre-training approaches. The error maps on the right highlight incorrect predictions, marked in red. For optimal viewing, please refer to the color version and zoom in for greater detail.
	}
	\label{fig:vis_results_sk_appendix}
\end{figure*}

In Figures~\ref{fig:vis_results_nuscenes_appendix}, and \ref{fig:vis_results_sk_appendix}, we provide additional qualitative results from the fine-tuning experiments on downstream tasks. The application of our pre-training strategies leads to a significant improvement in model performance compared to baselines initialized randomly. Particularly, our method outperforms SLiDR~\citep{sautier2022slidr}, demonstrating its enhanced ability to handle segmentation tasks. While our approach shows substantial improvements, we observe some false positives in challenging scenarios, which we plan to address in future work.

%\section{\minor{Detailed Discussion on 3D Pretext Tasks: MAE vs. Occupancy Estimation}}
\section*{D. Further Discussion}
%\section{\minor{Discussion on 3D Pretext Tasks: MAE vs. Occupancy Estimation}}

\minor{In Section 4.1, we selected occupancy estimation over Masked Autoencoders (MAE) for the 3D branch. While patch-based MAE methods (e.g., Point-MAE~\citep{pang2022masked}) are highly effective for dense, object-centric point clouds---since their FPS and KNN sampling ensures masked patches always contain real points---applying MAE to sparse outdoor scenes presents unique challenges.}

\minor{For large-scale sparse scenes, MAE typically requires voxelization. As Voxel-MAE~\citep{hess2023masked} highlights, masking only non-empty voxels creates an occupancy shortcut, allowing the model to trivially assume every masked token is a solid object. To prevent this, Voxel-MAE must also mask inherently empty voxels. However, this introduces ambiguity: the network must implicitly deduce whether a masked token represents physically empty space or a masked object before attempting reconstruction.}

\minor{Our framework avoids this indirect learning process. Occupancy estimation explicitly queries random spatial locations and directly supervises the network to predict their status (occupied vs. empty). This provides a clear, unambiguous signal for robustly learning 3D structural geometry in sparse environments.}

%\clearpage

\end{document}